\documentclass{article}

\usepackage{verbatim}

\usepackage[preprint,nonatbib]{neurips_2024} 

\usepackage[utf8]{inputenc} 
\usepackage[T1]{fontenc}    
\usepackage{hyperref}       
\usepackage{url}            
\usepackage{booktabs}       
\usepackage{amsfonts}       
\usepackage{nicefrac}       
\usepackage{microtype}      
\usepackage[dvipsnames]{xcolor}         
\usepackage{multirow}

\usepackage{tikz}
\usetikzlibrary{positioning}
\usepackage[british]{babel}

\usepackage{amsmath}
\usepackage{amssymb}
\usepackage{amsthm}
\usepackage{csquotes}

\usepackage{subcaption}

\usepackage[%
  giveninits  = true,
  maxbibnames = 99,
  hyperref    = true,
  backend     = biber,
  style       = numeric,
  sorting     = nty,
  sortcites,
]{biblatex}

\DefineBibliographyExtras{british}{}
\DeclareMathOperator*{\argmax}{arg\,max}

\usepackage{amssymb}
\usepackage{mathtools}
\usepackage{bbm}
\usepackage{amsthm}
\usepackage{cleveref}
\usepackage[inline]{enumitem}
\usepackage{wrapfig}
\usepackage{siunitx}
\usepackage{array}

\newcommand{\ruben}[1]{{\color{purple}#1}}

\renewcommand{\paragraph}[1]{{\vspace{0.6mm}\noindent \bf #1}.}

\newlist{inlinearabic}{enumerate*}{1}
\setlist*[inlinearabic,1]{%
  label=(\arabic*),
}
\theoremstyle{definition} 

\newtheorem{innerexample}{Example}[section]
\newenvironment{example}[1][]{%
  \if\relax\detokenize{#1}\relax
    \begin{innerexample}%
  \else
    \begin{innerexample}[#1]%
  \fi
}{%
  \end{innerexample}%
}

\title{Attending to Topological Spaces:\\ The Cellular Transformer}

\author{
  Rub\'en Ballester\\
  Departament de Matem\`atiques i Inform\`atica\\
  Universitat de Barcelona\\
  08007 Barcelona, Spain \\
\texttt{ruben.ballester@ub.edu} \\
  \and
  Pablo Hern\'andez-Garc\'{\i}a \\
  Departamento de Matem\'aticas\\
  Universidad de Salamanca\\
\texttt{pablohg.eka@usal.es} \\
  \and 
  Mathilde Papillon \\
  Department of Electrical Engineering\\
  University of California Santa Barbara\\
\texttt{papillon@ucsb.edu} \\
  \and
  Claudio Battiloro \\
  Department of Biostatistics\\
  Harvard University\\
\texttt{cbattiloro@hsph.harvard.edu} \\
  \and
  Nina Miolane \\
  Department of Electrical Engineering\\
  University of California Santa Barbara\\
\texttt{ninamiolane@ucsb.edu} \\
  \and
  Tolga Birdal \\
  Department of Computer Science\\
  Imperial College London\\
\texttt{t.birdal@imperial.ac.uk} \\
\and
  Carles Casacuberta \\
  Departament de Matem\`atiques i Inform\`atica\\
  Universitat de Barcelona\\
  08007 Barcelona, Spain \\
\texttt{carles.casacuberta@ub.edu} \\
\and
  Sergio Escalera \\
  Departament de Matem\`atiques i Inform\`atica\\
  Universitat de Barcelona\\
  08007 Barcelona, Spain \\
\texttt{sergio.escalera.guerrero@gmail.com} \\
  \and
  Mustafa Hajij \\
  Departament of Data Science\\
  University San Francisco\\
\texttt{mhajij@usfca.edu} \\
}

\begin{document}

\maketitle

\begin{abstract}
Topological Deep Learning seeks to enhance the predictive performance of neural network models by harnessing topological structures in input data. Topological neural networks operate on spaces such as cell complexes and hypergraphs, that can be seen as generalizations of graphs. In this work, we introduce the Cellular Transformer (CT), a novel architecture that generalizes graph-based transformers to cell complexes. First, we propose a new formulation of the usual self- and cross-attention mechanisms, tailored to leverage incidence relations in cell complexes, e.g., edge-face and node-edge relations. Additionally, we propose a set of topological positional encodings specifically designed for cell complexes. By transforming three graph datasets into cell complex datasets, our experiments reveal that CT not only achieves state-of-the-art performance, but it does so without the need for more complex enhancements such as virtual nodes, in-domain structural encodings, or graph rewiring.
\end{abstract}

\section{Introduction}
\label{sec:intro}
Topological Deep Learning (TDL) \autocite{papillon2023architectures, hajijtopological} is a fast growing field that leverages generalizations of graphs, such as simplicial complexes, cell complexes, and hypergraphs (collectively known as \textit{topological domains}) to extract comprehensive global information from data \autocite{bick2021higher, battiston2020networks}. Neural networks that are designed to process and learn from data supported on these topological domains form the core of TDL~\autocite{papillon2023architectures}. By transcending traditional graph-based representations, which are limited to binary relations in graph neural networks (GNNs), TDL exploits novel information contained in higher-order relationships, i.e., interactions involving multiple entities simultaneously. This capability provides new, unique opportunities to address innovative applications across diverse disciplines, including social sciences~\autocite{zhou2020graph}, transportation~\autocite{jiang2021graph}, physics~\autocite{battiston2021physics}, epidemiology~\autocite{deng2020cola}, as well as scientific visualization and discovery.

In parallel to TDL, the transformer architecture~\autocite{vaswani2017attention} has brought about a paradigm shift in learning on data with various modalities. Using multi-headed 
attention, transformers can capture long-range dependencies and hierarchical patterns in data like text or video~\autocite{dosovitskiy2020image, han2022survey}. Particularly, graph 
transformers~\autocite{wu2021representing, rong2020self} form a subset of transformers specifically designed to work with graph-based data. These models suffer from similar expressivity limitations as GNNs compared to Topological Neural Networks (TNNs), as they only represent pairwise relationships within the data.

\paragraph{Contributions} In this work, we propose to bridge the gap between TDL and transformers. We introduce the \emph{Cellular Transformer} (CT) to simultaneously harness the power of the expressive cell complex representation and the attention mechanism. By augmenting the transformer with topological awareness through cellular attention, CT is inherently capable of exploiting complex patterns in data mapped to high order representations, showing competitive or improved performance compared to graph and simplicial transformers and message passing architectures.

Our specific contibutions are summarized as follows.
\begin{enumerate}
[leftmargin=0.7cm, topsep=0.5pt]
    \item We propose the CT framework, which generalizes the graph-based transformer to process higher-order relations within cell complexes.
    \item We introduce cell complex positional encodings and formulate self-attention and cross-attention in topological terms. We demonstrate how to utilize these computational primitives to process data supported on cell complexes in a transformer layer.
    \item We benchmark the CT on three classical benchmark datasets, outperforming or achieving results comparable to the state-of-the-art without the need for complex enhancements of the architecture such as virtual nodes, involved in-domain structural or positional encodings or rewiring methods. 
\end{enumerate}



\section{Related Work}
\label{scn:related_work}

Transformer models have seen significant advancements in various domains, including natural language processing~\autocite{vaswani2017attention, kenton2019bert, radford2018improving}, computer vision~\autocite{dosovitskiy2020image,arnab2021vivit,han2022survey}, or graph learning.



\paragraph{Graph transformers}
Graph transformers are a special subset of transformers designed to learn from data supported on graphs. This evolving field encompasses three distinct strategies to harnessing the power of transformers in graph contexts. The first approach involves integrating GNNs directly into transformer architectures, either by stacking~\autocite{wu2021representing, rong2020self}, interweaving~\autocite{lin2021mesh}, or running in parallel~\autocite{zhang2020graph}. A second method focuses on encoding the graph structure into positional embeddings, which are then added to the input of the transformer model for spatial awareness. These embeddings can be computed in a myriad of ways, such as from Laplacian eigenvectors~\autocite{dwivedi2021generalization} or SVD vectors of the adjacency matrix~\autocite{hussain2021edge}. Finally, the third approach hard codes adjacency information into the self-attention. In~\autocite{dwivedi2021generalization, min2022masked}, only adjacent nodes are allowed to attend to each other, as all other attentional weights are zeroed. \autocite{mialon2021graphit} similarly manipulates self-attention via the kernel matrix instead of the adjacency matrix. We refer the reader to~\autocite{min2022transformer} for more information. Our work adopts a combination of the second and third methods in order to adapt the transformer to data supported on cell complexes.

\paragraph{Higher-order transformers}
Transformer models which go beyond pairwise relations represent a natural progression from graph-based transformers. The most prominent category of such higher order transformers operates on hypergraphs. In many instances, they have also adopted the self-attention mechanism \cite{kim2021transformers, hu2021semi, zhang2020hypersagnn, wang2020next}. Beyond graphs and hypergraphs, transformers operating on topological domains are scarce. To our knowledge, only two higher-order transformers operating on simplicial complexes have been proposed~\autocite{Clift2020Logic, zhou2024theoretical}. The first approach, however, does not consider higher order features directly, but rather leverages higher order relations to improve features on nodes. 
The second approach, although being a fairly general object to define higher-order structures, focuses primarily on graph learning. 
It proposes two architectures: one operating on tuples of nodes (i.e., cliques) within the graph, which may not naturally 
appear in the clique complex of the graph, and another applying a general attention mechanism to all simplices in a lifted graph simultaneously, disregarding the distinct nature of data across dimensions (e.g., properties of atoms in nodes vs. properties of bonds in edges). The latter was only tested on nodes and edges, excluding higher-order elements. A limitation of simplicial approaches is their representative power, as triangles and tetrahedra are scarce in natural data domains.


\paragraph{Non-transformer topological neural networks}
Besides transformers, 

recent years have witnessed a growing interest in higher-order networks~\autocite{battiston2020networks,bick2021higher}. In signal processing and deep learning, various approaches, such as Hodge-theoretic methods, message-passing schemes, and skip connections, have been developed for TNNs. The use of Hodge Laplacians for data analysis has been investigated in~\autocite{jiang2011statistical, lim2020hodge} and extended to a signal processing context, for example, in~\autocite{schaub2021signal,sardellitti2021topological,roddenberry2021signal} for simplicial and cell complexes. Convolutional operators and message-passing algorithms for TNNs have been developed. For hypergraphs, a convolutional operator has been proposed in~\autocite{jiang2019dynamic,feng2019hypergraph,arya2018exploiting} and has been further investigated in~\autocite{jiang2019dynamic,bai2021hypergraph,gao2020hypergraph}. Message passing on simplicial and cell complexes are proposed in \cite{hajijcell,hajij2021simplicial}. The expressive power of these networks is studied in \cite{CWNetworks}. Moreover, message passing on network sheaves can be found in \cite{hansen2019toward,hansen2020sheaf, sheaf2022, battiloro2022tangent, battiloro2023tangent, barbero2022sheaf}. A model able to infer a latent regular cell complex from data has been introduced in \cite{battiloro2024dcm}. Recently, message passing-free architectures have been introduced for simplicial complexes \cite{gurugubelli2024sann,maggs2024simplicial,ramamurthy2023topo}. 

Most attention-based models are designed primarily for graphs, with some recent exceptions that have been introduced in topological domains~\autocite{bai2021hypergraph,kim2020hypergraph,giusti2022simplicial,goh2022simplicial}. Attention on cell complexes was introduced in~\autocite{giusti2023cell} exploiting higher-order topological information through feature lifting and attention mechanisms over lower and upper neighborhoods, and a generalized attention mechanism on combinatorial complexes was introduced in~\autocite{hajijtopological,hajij2023combinatorial}. However, neither \autocite{giusti2023cell} nor \autocite{hajijtopological} consider query-key-value attention and positional/structural encodings. The work in  \autocite{giusti2023cell} works at the edge level and does not leverage any interplay among cells of different ranks. Furthermore, the work in \autocite{hajijtopological} does not account for dense attention, and, being based on the general notion of combinatorial complex, it is not able to readily encode and leverage specific topological features peculiar to cell complexes.

\section{Cell Complexes}
\label{cc:cell_complexes}

Cell complexes encompass various kinds of topological spaces used in network science, including graphs, simplicial complexes, and cubical complexes. A precise definition can be found in~\autocite{hatcher2005algebraic}, and further information is provided in the Appendix.


We constrain ourselves with $2$-dimensional regular cell complexes for simplicity, although our constructions and discussion carry over to higher-dimensional cellular complexes similarly. Thus, in this work, as in~\autocite{roddenberry2021signal}, a \textit{cell complex} is a triplet $\mathcal{X}=(\mathcal{X}_0, \mathcal{X}_1 , \mathcal{X}_2 )$ of finite ordered sets, where elements $v\in\mathcal{X}_0$ are called \emph{nodes}, \emph{vertices}, or \emph{$0$-cells}, elements $e\in\mathcal{X}_1$ are called \emph{edges} or \emph{$1$-cells}, and elements $\sigma\in\mathcal{X}_2$ are called \emph{faces} or \emph{$2$-cells}. Additionally, \emph{incidence relations} represent each edge as an ordered pair of vertices $e=[v_1,v_2]$ incident to $e$, and each face as an ordered sequence of edges $\sigma=[e_1,\dots,e_{m(\sigma)}]$ that form a closed path without self-intersections and constitute the edges incident to~$\sigma$. We assume that $m(\sigma)\ge 3$ to ensure that the pair $(\mathcal{X}_0,\mathcal{X}_1)$ is a (loopless, simple, directed) graph. The subscript $k$ of each set $\mathcal{X}_k$ is called its~\emph{rank}.

The incidence relations endow each edge and each face with an orientation.
For an edge $e=[v_1,v_2]$, the oppositely oriented edge is denoted by $-e = [v_2, v_1]$.
A collection of edges $e_1,\dots,e_m$ form a closed path if there is a set of distinct vertices $v_1,\dots,v_m$ such that $\pm e_i=[v_i,v_{i+1}]$ for~$1\le i\le m$ and $v_{m+1}=v_1$.
Incidence relations between cells of consecutive ranks are encoded into \emph{incidence matrices}. Thus, the first incidence matrix $\mathbf{B}_1$ has $(i,j)$ entry equal to~$-1$ if the $j$-th edge $e_j$ starts at the $i$-th vertex~$v_i$, $1$ if $e_j$ ends at~$v_i$, and $0$ otherwise. The entries of the second incidence matrix $\mathbf{B}_2$ are the incidence numbers between faces and edges, where the \emph{incidence number} of a face $\sigma$ with an edge $e$ is the sign of $\pm e$ if it belongs to a closed path of edges incident to~$\sigma$, and  $0$ otherwise. These two matrices satisfy $\mathbf{B}_1\mathbf{B}_2=0$, as shown in
\cite{hatcher2005algebraic,grady2010discrete}.
The \emph{non-signed incidence matrices} $\mathbf{I}_k$ for $k=1,2$, are obtained by replacing incidence numbers in $\mathbf{B}_k$ by their absolute values.

We refer to \emph{neighborhood matrices}, including signed and non-signed incidence matrices, upper and lower adjacency matrices, and Hodge Laplacians, as defined in detail in the Appendix.

\subsection{Data on cell complexes: cochain spaces}
\label{sec:data}
\begin{figure}[ht]
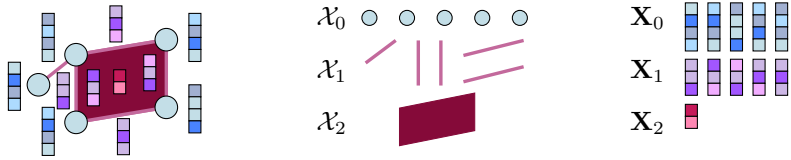

	\centering
        \include{assets/cell_complex}
        \vspace{-2em}

\caption{Illustration of an annotated cell complex. Left: An annotated cell complex $\mathcal{X}$ consisting of five vertices, five edges, and one $2$-cell.
Center: $\mathcal{X}_{k}$ is the collection of $k$-cells of~$\mathcal{X}$ for $k=0,1,2$. Right: Rows depict values of a cochain $\mathbf{X}_k$ for each~$k$, of dimensions $d_0=4$, $d_1=3$ and~$d_2=2$.\vspace{-4mm}}
\label{fig:cochains}
\end{figure}

Cochain spaces are used to process data supported over a cell complex $\mathcal{X}=(\mathcal{X}_0, \mathcal{X}_1 , \mathcal{X}_2)$. For $k=0,1,2$, we denote by $\mathcal{C}^k(\mathcal{X},\mathbb{R}^d )$
the $\mathbb{R}$-vector space of functions
$\mathcal{X}_k\to \mathbb{R}^d$, where $d\ge 1$. Here $d$ is called \emph{data dimension} and elements of $\mathcal{C}^k(\mathcal{X},\mathbb{R}^d)$ are called \emph{$k$-cochains} or \emph{$k$-signals} on~$\mathcal{X}$. For short, we write $\mathcal{C}^k(\mathcal{X})$ instead of~$\mathcal{C}^k(\mathcal{X},\mathbb{R})$ when $d=1$; see \cref{fig:cochains} for an example. 

An \emph{annotated cell complex} is a cell complex $\mathcal{X}$ together with a $k$-cochain $\mathbf{X}_k$ of dimension $d_k$ for each rank~$k$.
We view $\mathbf{X}_k$ as a matrix in $\mathcal{M}(|\mathcal{X}_k|,d_k)$, that is, with $|\mathcal{X}_k|$ rows and $d_k$ columns, whose $i$-th row is the image of the $i$-th element of $\mathcal{X}_k$.
In this work, all datasets consist of annotated cell complexes sharing the same dimensions $d_0$, $d_1$ and~$d_2$.

\section{The Cellular Transformer}
\label{scn:cellular_transformer}
In this section, we present a general transformer architecture for cell complexes. First, we discuss different approaches to perform attention on cells and define the cellular transformer layer. Then, we propose different positional encoding methods for the cellular transformer, that identify cells by their relative position in the cell complex or by their \emph{centrality} according to random walks, as in~\autocite{dwivedi2022graph}.

\subsection{Overview}
\label{general}
\label{CT_spaces}
A \emph{cellular transformer} is a neural network which, given an annotated cell complex $\mathcal{X}$, induces a composition of functions $\text{CT} = \text{R}\circ \text{CT}_{L}\circ \cdots \circ \text{CT}_{1}\circ\text{P}$, named layers, where $\text{P}$ is a preprocessing layer, as described in~\cref{scn:high_order_positional_encodings}, $\text{R}$ is a readout layer that converts cochains on top of cells into an output prediction value, and $\text{CT}_l$, for $l=1,\hdots, L$, are cellular transformer layers defined as functions of the form
\begin{equation}
\label{eqn:domain_ct_layer}
\text{CT}_l\colon \mathcal{C}^{0}(\mathcal{X}, \mathbb{R}^{d^h_{0}}) \times \cdots \times \mathcal{C}^{n}(\mathcal{X}, \mathbb{R}^{d^h_{n}}) \longrightarrow \mathcal{C}^{0}(\mathcal{X}, \mathbb{R}^{d^h_{0}}) \times \cdots \times \mathcal{C}^{n}(\mathcal{X}, \mathbb{R}^{d^h_{n}}),
\end{equation}
where $n=\dim\mathcal{X}$ and $h$ indicates the dimension of hidden layers. In our experiments, we set $d_0^h = \cdots = d_n^h$, whose value depends on the dataset and is specified in the Appendix.

\Cref{eqn:domain_ct_layer} does not provide an explicitly parametrization of the cellular transformer layer as it only describes the function (co)domains. A parametrization of the cellular transformer layers can be given using tensor diagrams together with the cellular attention formulae, described in~\cref{scn:attention_arch}. Transformer and preprocessing layers take as input an annotated cell complex and output the same cell complex with different cochains. We denote input $k$-cochains on the $\text{CT}_l$ layer as~$\mathbf{X}_{k, l}$.


\subsection{The cellular attention layer and cellular transformer architecture}
\label{scn:attention_arch}

We propose two cellular attention mechanisms for transformer layers. The first mechanism generalizes self- and cross-attention and depends on the dimensions of the cells. We call it \emph{pairwise cellular attention}. The second mechanism performs the attention over all cells ignoring their dimension. We call it \textit{general cellular attention}. We discuss in~\cref{sec:discussion} the data regimes in which each attention mechanism performs better than the other.



\subsubsection{Pairwise cellular attention}
\label{scn:cell_self_cross_att}
Our first mechanism performs pairwise attention between cells of arbitrary ranks according to a tensor diagram (see~\cref{scn:tensor_diagrams_for_ct}), and then aggregates the outputs received for the same rank. Given source and target ranks $0\leq k_s, k_t \leq \dim \mathcal{X}$ and cochains $\mathbf{X}_{k_t} $, $\mathbf{X}_{k_s}$, the single-head attention from $k_s$ to $k_t$ is a map $\mathcal{C}^{k_s}(\mathcal{X}, \mathbb{R}^{d^h_s})\times \mathcal{C}^{k_t}(\mathcal{X}, \mathbb{R}^{d^h_t}) \to \mathcal{C}^{k_t}(\mathcal{X}, \mathbb{R}^{d^h_t})$ defined as 
\begin{equation}
\label{formula:attention_self_cross}
\mathcal{A}_{k_s\to k_t}^\bullet(\mathbf{X}_{k_t}, \mathbf{X}_{k_s})= \text{softmax}(\mathbf{X}_{k_t}\mathbf{Q}_{k_s\to k_t}(\mathbf{X}_{k_s}\mathbf{K}_{k_s\to k_t})^T \star \phi(\mathbf{N}_{k_s\to k_t})) \mathbf{X}_{k_s}\mathbf{V}_{k_s\to k_t}, 
\end{equation}
where $\mathbf{Q}_{k_s\to k_t}\in\mathcal{M}(d^h_t, p)$, $\mathbf{K}_{k_s\to k_t}\in \mathcal{M}(d^h_s, p)$, and $\mathbf{V}_{k_s\to k_t}\in \mathcal{M}(d_s^h, d^h_t)$ are learnable query, key, and value real matrices with $p$ a fixed hyperparameter shared by all transformer layers.
 The symbol $\bullet\in\{d,s,c\}$ indicates whether we are performing dense, sparse, or a mixed type of attention, performing dense attention for cells of the same rank and sparse attention otherwise, respectively. The symbol $\star$ is a sum or a Hadamard product for dense or sparse attention, respectively. $\mathbf{N}$ is a neighborhood matrix, and $\phi$ is a function,
possibly with learnable parameters. For our experiments, we set $\phi(\mathbf{N})=\theta\mathbf{N}$ where $\theta$ is a learnable parameter for dense attention and the identity matrix for sparse attention. 
Attention formulae performs query, key, and value projections without bias for simplicity. A bias term can be added to the projections, as in most transformer implementations. Multi-head attention can also be performed by 
\begin{inlinearabic}
    \item splitting the cochains $\mathbf{X}_{k_s}$ and $\mathbf{X}_{k_t}$ into multiple cochains $\mathbf{X}_{k_s}^1, \hdots, \mathbf{X}_{k_s}^m$ and $\mathbf{X}_{k_t}^1, \hdots, \mathbf{X}_{k_t}^m$ of smaller dimension;
    \item performing single-head attention for each pair of cochains $\mathbf{X}_{k_s}^i, \mathbf{X}_{k_t}^i$;
    \item concatenating the 
    outputs of the single-head attention for the different pairs 
    into a full cochain of dimension $d^h_t$.
\end{inlinearabic}

For a specific rank $k_t$, CT layers can produce multiple attention outputs from different rank sources $k_s$. In the CT layer, we adopt the standard prenorm design~\autocite{on_layer_normalization}, where for each rank $k_t$, the outputs from the various rank sources $k_s$ are summed in the residual connection.
The specific 
algorithm for the CT layer is detailed in the Appendix.
In our experiments, we set $\mathbf{N}_{k\to k}$ to be the upper adjacency matrix for $k=0$ and the lower adjacency matrix if $k>0$, and $\mathbf{N}_{k_s\to k_t}$ to be the non-signed incidence matrix between dimensions $k_s$ and $k_t$ if $k_s > k_t$, and the transpose matrix otherwise.

\subsubsection{General cellular attention}
\label{scn:general_attention}
The second mechanism performs attention with all the cells at the same time, disregarding their rank, as proposed in~\autocite{zhou2024theoretical}. In the general attention mechanism, cells share the same key and query 
matrices, but have different value 
matrices for each rank. The single-head attention formula is
\begin{equation*}
\mathcal{A}_g^\bullet(\mathbf{X}) = \text{softmax}\left(\mathbf{X}Q(\mathbf{X}K)^T \star \phi(\mathbf N)\right)\left(\text{Concat}\left[(\mathbf{X}_0V^0)^T,\hdots, (\mathbf{X}_{\dim \mathcal{X}}V^{\dim \mathcal{X}})^T\right]\right)^T,
\end{equation*}
using the same notations as in the previous subsection. In this case, the prenorm transformer layer is performed as usual. The 
algorithm 
corresponding to the general attention CT layer is detailed in the Appendix. As in~\cref{scn:cell_self_cross_att}, multi-head attention can be performed by splitting the original cochains into smaller cochains, applying the general single-head attention to pairs of the smaller cochains, and concatenating again into a single, big cochain. In our experiments, we let $\mathbf{N}$ be the following combination of the previous $\mathbf{N}_{k_s \to k_t}$ matrices, where $n=\dim\mathcal{X}$:
\begin{equation}
\label{eqn:big_n_matrix_general_attention}
    \mathbf{N} =\begin{bmatrix}
\mathbf{N}_{0\to 0} & \mathbf{N}_{1\to 0} & \hdots & 0 \\
\mathbf{N}_{1\to 0}^T & \mathbf{N}_{1\to 1} & \hdots & 0 \\
\vdots & \vdots & \ddots &  \vdots\\
0 & 0 &\hdots  & \mathbf{N}_{n\to n} 
\end{bmatrix}.
\end{equation}

\subsection{Tensor diagrams for cellular transformers}
\label{scn:tensor_diagrams_for_ct}

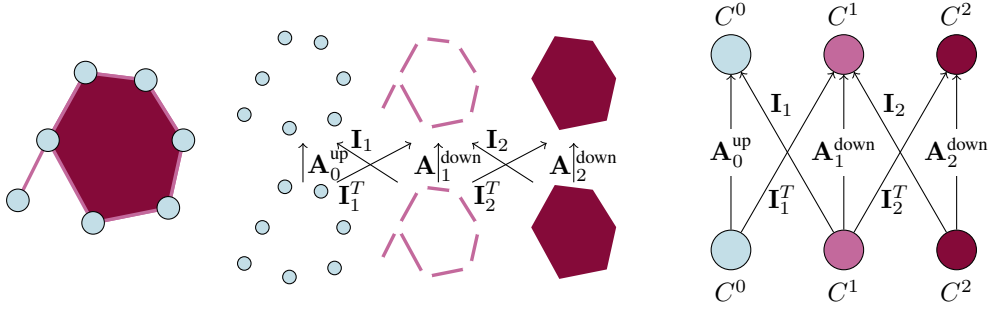
\begin{figure}[ht]
	\centering
        \begin{tikzpicture}[scale=0.90]


    \definecolor{nodecolor}{HTML}{c3dee7}
    \definecolor{edgecolor}{HTML}{c3699c}
    \definecolor{shapecolor}{HTML}{850a39}

    \definecolor{feturenode1}{HTML}{6394ff}
    \definecolor{featurenode2}{HTML}{ceddff}
    \definecolor{featurenode3}{HTML}{1c63ff}

    \definecolor{featureedge1}{HTML}{a963ff}
    \definecolor{featureedge2}{HTML}{bf8eff}
    
    \def\edgewidth{0.45mm}
    \def\noderadius{0.15}

\tikzset{
        featurenodes/.pic={
    \def\squaresize{0.16} 
    \def\gap{0.0} 

    \draw[fill=feturenode1, draw=black] (0, 0) rectangle (\squaresize, -\squaresize);
    \draw[fill=featurenode2, draw=black] (0, -\squaresize - \gap) rectangle (\squaresize, -2*\squaresize - \gap);
    \draw[fill=featurenode3, draw=black] (0, -2*\squaresize - 2*\gap) rectangle (\squaresize, -3*\squaresize - 2*\gap);
        }
}
\tikzset{
        featureedges/.pic={
    \def\squaresize{0.16} 
    \def\gap{0.0} 

    \draw[fill=featureedge1, draw=black] (0, 0) rectangle (\squaresize, -\squaresize);
    \draw[fill=featureedge2, draw=black] (0, -\squaresize - \gap) rectangle (\squaresize, -2*\squaresize - \gap);
        }
}
    \tikzset{
        hexagoncell/.pic={
            \coordinate (A) at (0, 0); 
            \coordinate (B) at (1, 0.2); 
            \coordinate (C) at (1.2, 1.1);
            \coordinate (D) at (0.7, 1.9);
            \coordinate (E) at (-0.1, 2.0);
            \coordinate (F) at (-0.6, 1.1);
            \coordinate (G) at (-1, 0.3);

            \draw[line width=\edgewidth, edgecolor] (A) -- (B);
            \draw[line width=\edgewidth, edgecolor] (B) -- (C);
            \draw[line width=\edgewidth, edgecolor] (C) -- (D);
            \draw[line width=\edgewidth, edgecolor] (D) -- (E);
            \draw[line width=\edgewidth, edgecolor] (E) -- (A);
            \draw[line width=\edgewidth, edgecolor] (E) -- (F);
            \draw[line width=\edgewidth, edgecolor] (F) -- (G);

            \filldraw[fill=shapecolor, draw=edgecolor, line width=\edgewidth] (A) -- (B) -- (C) -- (D) -- (E) -- (F) -- cycle;

            \foreach \point in {A, B, C, D, E, F, G} {
                \filldraw[fill=nodecolor, draw=black] (\point) circle (\noderadius);
            }
        }
    }

        \tikzset{
        hexagoncelldim0/.pic={
            \coordinate (A) at (0, 0); 
            \coordinate (B) at (1, 0.2); 
            \coordinate (C) at (1.2, 1.1);
            \coordinate (D) at (0.7, 1.9);
            \coordinate (E) at (-0.1, 2.0);
            \coordinate (F) at (-0.6, 1.1);
            \coordinate (G) at (-1, 0.3);



            \foreach \point in {A, B, C, D, E, F, G} {
                \filldraw[fill=nodecolor, draw=black] (\point) circle (\noderadius);
            }
        }
    }

            \tikzset{
        hexagoncelldim1/.pic={
            \coordinate (A) at (0, 0); 
            \coordinate (B) at (1, 0.2); 
            \coordinate (C) at (1.2, 1.1);
            \coordinate (D) at (0.7, 1.9);
            \coordinate (E) at (-0.1, 2.0);
            \coordinate (F) at (-0.6, 1.1);
            \coordinate (G) at (-1, 0.3);

            \draw[line width=\edgewidth, edgecolor] (A) -- (B);
            \draw[line width=\edgewidth, edgecolor] (B) -- (C);
            \draw[line width=\edgewidth, edgecolor] (C) -- (D);
            \draw[line width=\edgewidth, edgecolor] (D) -- (E);
            \draw[line width=\edgewidth, edgecolor] (F) -- (A);
            \draw[line width=\edgewidth, edgecolor] (E) -- (F);
            \draw[line width=\edgewidth, edgecolor] (F) -- (G);


            \foreach \point in {A, B, C, D, E, F, G} {
                \filldraw[fill=white, draw=white] (\point) circle (\noderadius);
            }
        }
    }

            \tikzset{
        hexagoncelldim2/.pic={
            \coordinate (A) at (0, 0); 
            \coordinate (B) at (1, 0.2); 
            \coordinate (C) at (1.2, 1.1);
            \coordinate (D) at (0.7, 1.9);
            \coordinate (E) at (-0.1, 2.0);
            \coordinate (F) at (-0.6, 1.1);
            \coordinate (G) at (-1, 0.3);


            \filldraw[fill=shapecolor, draw=shapecolor, line width=\edgewidth] (A) -- (B) -- (C) -- (D) -- (E) -- (F) -- cycle;

        }
    }

    \tikzstyle{C0} = [circle, fill=nodecolor, minimum size=15pt, inner sep=0pt, draw]
    \tikzstyle{C1} = [circle, fill=edgecolor, minimum size=15pt, inner sep=0pt, draw]
    \tikzstyle{C2} = [circle, fill=shapecolor, minimum size=15pt, inner sep=0pt, draw]
    
\tikzset{
        tensordiag/.pic={
    \node[C0, label=below:$C^0$] (C0_1) at (2, 0) {};
    \node[C0, label=above:$C^0$] (C0_2) at (2, 2.6) {};
    
    \node[C1, label=below:$C^1$] (C1_1) at (3.5, 0) {};
    \node[C1, label=above:$C^1$] (C1_2) at (3.5, 2.6) {};
    
    \node[C2, label=below:$C^2$] (C2_1) at (5, 0) {};
    \node[C2, label=above:$C^2$] (C2_2) at (5, 2.6) {};

    \draw[->] (C0_1) -- node[midway, fill=white] {$\mathbf{A}^{\text{up}}_0$} (C0_2);
    \draw[->] (C0_1) -- node[midway, yshift=-18pt, xshift=-2pt] {$\mathbf{I}_1^T$} (C1_2);
    \draw[->] (C1_1) -- node[midway, fill=white] {$\mathbf{A}^{\text{down}}_1$} (C1_2);
    \draw[->] (C1_1) -- node[midway, yshift=-18pt, xshift=-2pt] {$\mathbf{I}_2^T$} (C2_2);
    \draw[->] (C1_1) -- node[midway, yshift=18pt, xshift=-2pt] {$\mathbf{I}_1$} (C0_2);
    \draw[->] (C2_1) -- node[midway, fill=white] {$\mathbf{A}^{\text{down}}_2$} (C2_2);
    \draw[->] (C2_1) -- node[midway, yshift=18pt, xshift=-2pt] {$\mathbf{I}_2$} (C1_2);
        }
}

    \pic at (-0.4, -1.4) {hexagoncell};
    

    \pic[scale=0.6, name=lefttop] at (2.5, 0) {hexagoncelldim0};
    \pic[scale=0.6, name=centertop] at (4.5, 0) {hexagoncelldim1};
    \pic[scale=0.6, name=righttop] at (6.5, 0) {hexagoncelldim2};

    \pic[scale=0.6, name=leftdown] at (2.5, -2.2) {hexagoncelldim0};
    \pic[scale=0.6, name=centerdown] at (4.5, -2.2) {hexagoncelldim1};
    \pic[scale=0.6, name=rightdown] at (6.5, -2.2) {hexagoncelldim2};

    \draw[->] (2.7, -0.8) -- node[midway, right, xshift=-2pt, yshift=-1pt] {$\mathbf{A}^{\text{up}}_0$} (2.7, -0.22);
    \draw[->] (4.7, -0.8) -- node[midway, right, xshift=-13pt, yshift=-1pt] {$\mathbf{A\,}^{\text{down}}_1$}(4.7, -0.22);
    \draw[->] (6.7, -0.8) -- node[midway, right, xshift=-13pt, yshift=-1pt] {$\mathbf{A\,}^{\text{down}}_2$} (6.7, -0.22);

    \draw[->] (3.2, -0.8) -- node[near start, below right, xshift=-10pt] {$\mathbf{I}_1^T$} (4.3, -0.22);
    \draw[->] (4.1, -0.8) -- node[near end, above right, yshift=-3pt, xshift=-4pt] {$\mathbf{I}_1$} (3.2, -0.22);

    \draw[->] (5.2, -0.8) -- node[near start, below right, xshift=-10pt] {$\mathbf{I}_2^T$} (6.3, -0.22);
    \draw[->] (6.1, -0.8) -- node[near end, above right, yshift=-3pt, xshift=-4pt] {$\mathbf{I}_2$} (5.2, -0.22);
    
    \pic at (6.8, -1.8) {tensordiag};

\end{tikzpicture}
        
\vspace*{-0.8cm} 

\caption{Tensor diagram illustrating the flow of signals between cochains defined on $0$-, $1$-, and $2$-cells.
For pairwise attention (\cref{scn:cell_self_cross_att}), the neighborhood matrices indicate the bias $\mathbf{N}$ in the attention formula 
\eqref{formula:attention_self_cross}. 
For general attention (\cref{scn:general_attention}), neighborhood matrices indicates how to build the bias matrix $\mathbf{N}$ by composition of smaller bias matrices $\mathbf{N}_{k_s\to k_t}$ between dimensions.\vspace{-4mm}}
	\label{fig:td}
\end{figure}

Cellular transformers involve interactions between cochains of different ranks. Tensor diagrams~\autocite{hajijtopological} provide a graphical abstraction that illustrates the flow of information of one CT layer. 
A tensor diagram portrays a CT Layer through the use of a directed graph. Nodes of a tensor diagram represent cochain spaces for different ranks $0\leq k \leq n$, where $n$ is the maximum allowed rank of cell complexes processed by 
the CT layer. If the input cell complex $\mathcal{X}$ is of lower dimension than $n$, the attention on ranks 
$k>\dim \mathcal{X}$ 
are ignored. In turn, edges represent either the pairwise attentions performed in the CT layer together with the bias matrices $\mathbf{N}_{k_s\to k_t}$, 
or simply the matrices used to build the 
matrix $\mathbf{N}$ from smaller matrices $\mathbf{N}_{k_s\to k_t}$ as in~\cref{eqn:big_n_matrix_general_attention}, for the general attention. A missing arrow from cochains of rank $k_s$ to cochains of rank $k_t$ implies a zero in the block of $\mathbf{N}$ corresponding to the matrix $\mathbf{N}_{k_s\to k_t}$. An illustration of the tensor diagram used in our experiments 
is given in \cref{fig:td}.

\vspace*{-0.3cm}

\subsection{Positional encodings on cellular complexes}
\label{scn:high_order_positional_encodings}
Transformers do not leverage the input structure explicitly by default~\autocite{vaswani2017attention}. Positional encodings help to overcome this problem by injecting positional and structural information about the input \emph{tokens}. For sequences, the first positional encoding used sine and cosine functions depending on the position of the token in the sequence. For graphs, several positional encodings have been studied such as the eigenvectors of the graph Laplacian (LapPE)~\autocite{benchmarking_gnn} and Random Walk Positional Encodings (RWPe)~\autocite{dwivedi2022graph}, where the latter were also adapted for simplicial complex transformers~\autocite{zhou2023facilitating, high_order_random_walks}. 


Let  $0\leq k \leq \dim \mathcal{X}$, where $\mathcal{X}$ is a cell complex. A cellular $k$-positional encoding of $\mathcal{X}_k$ is a $k$-cochain $\mathbf{E}_k$ that captures some structural information about $\mathcal{X}_k$ within $\mathcal{X}$ (positional encoding may also be defined on the entire complex~$\mathcal{X}$). 
Given cochains $\mathbf{X}_{k}$ and positional encodings $\mathbf{E}_k$, the input for the first transformer layer is defined as a function $\text{P}_k\colon \mathcal{C}^k(\mathcal{X}, \mathbb R^{d_k})\times \mathcal{C}^k(\mathcal{X}, \mathbb R^{d_\text{pe}})\to \mathcal{C}^k(\mathcal{X}, \mathbb R^{d^h_k})$ with $
    \mathbf{X}_{k}^1 = \text{P}_k(\mathbf{X}_{k}, \mathbf{E}_k)$,
where $\text{P}_k$ combines the signals and the positional encodings. Usual functions are
\begin{equation}
\begin{split}
    \text{SumPE}(\mathbf{X}_{k}, \mathbf{E}_k) &= \mathbf{X}_{k}\theta_{\text{in}, k} + b_{\text{in}, k} + \mathbf{E}_k \theta_{\text{in}, \text{pe}} + b_{\text{in}, \text{pe}} \\
    \text{ConcatPE}(\mathbf{X}_{k}, \mathbf{E}_k) &= \text{Concat}(\mathbf{X}_{k}, \mathbf{E}_k)\theta_{\text{in}, \text{pe}} + b_{\text{in}, \text{pe}},
    \end{split}
\end{equation}
where $\theta_\bullet$, $b_\bullet\in$ are learnable parameters. For this paper, we use $\text{P}_k=\text{ConcatPE}$. 

Next we discuss three novel  positional encodings on cell complexes \ref{scn:LapPE}: Barycentric Subdivision 
\ref{sec:randomwalk}, Random Walk, 
and Topological Slepians 
\ref{sec:Slepians}. 


\subsubsection{BSPe: Barycentric Subdivision Positional Encoding}
\label{scn:LapPE}
A popular positional encoding for graph transformers is given by graph Laplacian eigenvectors (LapPE)~\autocite{benchmarking_gnn}. 
LapPE assigns to each vertex $v_i$ a vector $\text{LapPE}(v_i)=(e^1_{i},\hdots, e^k_{i})$, where $\{e_i^j\mid j=1,\dots,k\}$
are eigenvectors of the 
$k$ smallest eigenvalues of the normalized graph Laplacian for a graph $G=(V,E)$, counting multiplicities, where $k$ is a hyperparameter. 

We denote the naive extension from LapPE for cell complexes using the unnormalized Hodge Laplacian matrix, instead of the graph Laplacian one, as \textbf{HodgeLapPE}. We use the unnormalized version because normalizing the Hodge Laplacian for dimensions greater than zero is not an easy task~\autocite{high_order_random_walks}. HodgeLapPE are, however, not a good choice for high-order positional encodings \emph{a priori} due to both a lack of normalization and the 
ambiguous information contained in Hodge Laplacians for nonzero rank. Details on LapPE for graphs and their HodgeLapPE extension are in the Appendix.

To overcome the previous drawbacks, we propose to extend LapPE by taking the original Laplacian positional encodings of the $1$-skeleton of the barycentric subdivisions of the cell complexes. 
The \emph{barycentric subdivision} of a cell complex $\mathcal{X}$, denoted by $\Delta(\mathcal{X})$, is the order complex of its face poset \autocite{wachs2006poset}, i.e., the abstract simplicial complex whose set of vertices is the set of cells of $\mathcal{X}$ and whose simplices are the totally ordered flags of cells of~$\mathcal{X}$. 
Barycentric subdivisions yield triangulations of cell complexes that preserve their topological properties
\autocite{cooke1967homologyofcellcomplexes}.
The $1$-skeleton of the barycentric subdivision of $\mathcal{X}$ is a graph $G=(V, E)$ where $V$ is the set of cells of $\mathcal{X}$ and where two vertices $\sigma_1$ and $\sigma_2$ are connected if one is a face of the other. 
The positional encoding of a cell $\sigma$ is the Laplacian positional encoding of $\sigma$ seen as a vertex in~$G$. 

\vspace*{-0.2cm}

\begin{figure}[htb!]
    \centering
    \begin{subfigure}[t]{0.3\textwidth}
    \centering
     \includegraphics[width=0.8\textwidth]{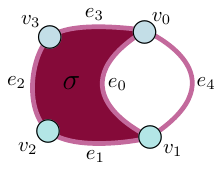}
    \end{subfigure}
    \hfill
     \begin{subfigure}[t]{0.3\textwidth}
     \centering
     \includegraphics[width=0.85\textwidth]{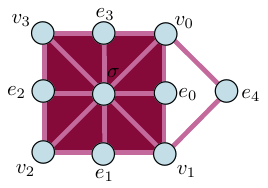}
    \end{subfigure}
    \hfill
     \begin{subfigure}[t]{0.3\textwidth}
     \centering
     \includegraphics[width=0.85\textwidth]{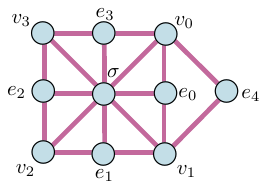}
    \end{subfigure}
    \caption{Left: A cell complex $\mathcal{X}$. Center: Barycentric subdivision of $\mathcal{X}$. Right: $1$-skeleton of the barycentric subdivision. Each original cell of $\mathcal{X}$ is represented by a node in the $1$-skeleton.}
    \label{fig:BarycSubdivision}
\end{figure}
This positional encoding respects the same theoretical advantages of the LapPE while assigning relative positions to all the cells \emph{at the same time}, and thus relative positions take into account all the cells and not only the cells of a specific dimension. We say that the positional encodings satisfying this property are called \emph{global}, in contrast to \emph{local} positional encodings, where encodings are assigned independently for each dimension. We denote this positional encoding as \textbf{BSPe}.

\subsubsection{RWPe: Random Walk Positional Encoding}
\label{sec:randomwalk}
RWPe take another different way of positioning vertices on a graph based on random walks. Given a vertex $v_i\in V$, the RWPe of $v_i$ is given by the vector $\text{RWPe}(v_i)=\left(\text{RW}_{ii},\hdots, \text{RW}_{ii}^k\right)$ where $\text{RW}=AD^{-1}$ is the random walk operator of a graph based on edge connectivity. In this case, each vertex is assigned the probabilities of landing again on itself on the random walks from one to $k$ steps. In this case, $\text{RWPe}(v_i)$ is unique and does not need sign or eigenvector selection invariance. 
For RWPe we have again difficulties defining meaningful random walks on general cell complexes. This problem is first explored in~\autocite{high_order_random_walks} and then in~\autocite{zhou2023facilitating} for simplicial complexes, making a special focus on edge random walks. There, transition matrices are defined to perform random walks in simplices of an arbitrary rank and are not affected by the orientation of the simplicial complex.

As a first, more naive approach, we propose to extend RWPe to cell complexes by taking the positional encodings given by the original RWPe for the $1$-skeleton of the barycentric subdivision introduced in~\cref{scn:LapPE}. We denote these \emph{global} positional encodings as \textbf{RWBSPe}. We also propose a more sophisticated, local approach, denoted \textbf{RWPe}, extending the random walks from~\autocite{high_order_random_walks} to cell complexes. The full development of the random walk matrix can be found in the Appendix. From the random walk matrix, positional encodings are taken as in RWPe for each rank of the cell complex.

\subsubsection{TopoSlepiansPE: Topological Slepians Positional Encoding}
\label{sec:Slepians}
The objective of positional encoding is to assign a relative position to tokens informed by the underlying domain, which for us are cells within a cell complex. Topological Slepians, as introduced in~\autocite{topo_slepians}, represent a novel category of signals specifically defined over cell complexes. These signals are characterized by their maximal concentration within the topological domain (the cells) and their perfect localization in the corresponding dual domain (the frequencies ~\autocite{sardellitti2022topological}). This means that the non-zero coordinates of the different topological Slepians are concentrated only on some of the cells of the complex, with a specific spectral content. 
Topological Slepians have been shown to be particularly efficient for signal representation and sparse coding tasks~\autocite{topo_slepians}, making them a valuable tool to devise positional encoding. 

In this work, we compute Slepians as in Section~4 of~\autocite{topo_slepians}.
Given an ordered multiset of rank~$k$ Slepians $\mathcal{S}_k=\{s_k^i\mid i=1,\dots,d_{\text{pe}}\}$, 
we let the positional encoding of a $k$-cell $\sigma$ be 
$\mathbf{E}(\sigma)_i=s_{k, \sigma}^i$,
the coordinate corresponding to $\sigma$ of the $s_k^i$ Slepian (in practice, we can obtain $\left|\mathcal{S}_k\right|=d_{\text{pe}}$ by adding zero vectors). We denote this positional encoding as \textbf{TopoSlepiansPE}. Although it comes with guaranteed localization properties, it is a local positional encoding, and suffers from permutation variance and eigenvector sign variance, that the transformers need to learn.

\section{Experiments}
\label{scn:experiments}

\begin{table}[t]
    \centering
    \caption{Models and scores compared for the 
    datasets \texttt{GCB}, \texttt{ZINC}, and \texttt{ogbg-molhiv}. Average and standard deviation for accuracy ($\uparrow$) are shown for the \texttt{GCB} test dataset; MAE ($\downarrow$) is reported for the \texttt{ZINC} test dataset; AUC-ROC ($\uparrow$) is reported for the \texttt{ogbg-molhiv} test dataset. \texttt{GCB} models are described in the Appendix. Results for \texttt{GCB} were extracted from~\autocite{gcb_repository}, and other results were extracted from~\autocite{zhou2024theoretical}. The first seven model rows   represent message passing architectures. For the \texttt{GCB} dataset, the other six models are classic machine learning algorithms. For the \texttt{ZINC} and \texttt{OGB} datasets, the second set of rows correspond to graph transformer models, and the third set of rows belong to the simplicial transformer models
of~\autocite{zhou2024theoretical}. The abbreviation v.n. means virtual node.}
    \vspace{3pt}
    \label{tab:summary_all_results}
    \begingroup
    \setlength{\tabcolsep}{2pt}
    \begin{tabular}[t]{lclclc}
        \toprule
        \multicolumn{2}{c}{\texttt{GCB}}                       & \multicolumn{2}{c}{\texttt{ZINC}} & \multicolumn{2}{c}{\texttt{ogbg-molhiv}}                                                                                       \\
        \cmidrule(lr){1-2}\cmidrule(lr){3-4}\cmidrule(lr){5-6}
        Model                                                       & Accuracy ($\uparrow$)             & Model                                    & MAE ($\downarrow$) & Model                                   & AUC-ROC ($\uparrow$) \\
        \midrule
        Graclus~\autocite{graclus}                                  & $0.690\pm 0.015$                  & GCN~\autocite{GCN}                       & $0.367$            & GCN+v.n.                                & $0.7599$             \\
        NDP~\autocite{ndp_model}                                    & $0.726 \pm 0.009$                 & GAT~\autocite{GAT}                       & $0.384$            & GIN+v.n.~\autocite{xu2018powerful}\hspace{-1.1em}                                & $0.7707$             \\
        DiffPool~\autocite{diffpool_model}                          & $0.699 \pm 0.019$                 & GatedGCN~\autocite{ResidualGatedGCN}     & $0.282$            & DGN~\autocite{DGN}                      & $0.7970$             \\
        Top-K~\autocite{graphunets}                                 & $0.427 \pm 0.152$                 & PNA~\autocite{PNA}                       & $0.188$            & PNA                                     & $0.7905$             \\
        SAGPool~\autocite{pmlr-v97-lee19c}                          & $0.377 \pm 0.145$                 &                                          &                    & GSN~\autocite{GSN}                      & $0.8039$             \\
        MinCutPool~\autocite{bianchi2020spectral}                   & $0.738 \pm 0.019$                 & CIN~\autocite{CWNetworks}                & $0.079$            & CIN                                     & $\mathbf{0.8094}$    \\
        ESC + RBF-SVM~\autocite{graph_embedding_and_classification} & $0.625 \pm 0.046$                 & GIN-AK+~\autocite{GNNAK}                 & $0.080$            & GIN-AK+                                 & $0.7961$             \\
        \cmidrule(lr){1-6}
        ESC + L1-SVM~\autocite{graph_embedding_and_classification}  & $0.722 \pm 0.010$                 & Graphormer~\autocite{Graphormer}         & $0.122$            &                                         &                      \\
        ESC + L2-SVM~\autocite{graph_embedding_and_classification}  & $0.693 \pm 0.016$                 & SAN~\autocite{RethinkingGTLap}           & $0.139$            & SAN                                     & $0.7785$             \\
        Hist Kernel~\autocite{hypergraph_kernels_over_sc}           & $0.720 \pm 0.000$                 & EGT~\autocite{EGT}                       & $0.108$            &                                         &                      \\
        Jaccard Kernel~\autocite{hypergraph_kernels_over_sc}        & $0.630 \pm 0.000$                 & GPS~\autocite{GPS}                       & $\mathbf{0.070}$            & GPS                                     & $0.7880$             \\
        \cmidrule(lr){3-6}
        Edit Kernel~\autocite{hypergraph_kernels_over_sc}           & $0.600 \pm 0.000$                 & $\mathcal {AS}_{0:1}^{\mathsf {SN}}$     & $0.080$            &                                         &                      \\
        Stratedit Kernel~\autocite{hypergraph_kernels_over_sc}      & $0.600 \pm 0.000$                 & $\mathcal {AS}_{0:1}^{\mathsf {SN+VS}}$~\autocite{zhou2024theoretical}  & $0.073$            & $\mathcal {AS}_{0:1}^{\mathsf {SN+VS}}$ & $0.7981$             \\
        \midrule
        $\mathcal {C}$ (ours)                                       & $\mathbf{0.752 \pm 0.010}$        & $\mathcal {C}$ (ours)                    & $0.080$            & $\mathcal {C}$ (ours)                   & $0.7946$             \\
        \bottomrule
    \end{tabular}\vspace{-4mm}
    \endgroup
    \label{tab:summary_all_experiments}
\end{table}

\begin{table}[t]
    \centering
    \caption{Table containing results for the best attention mechanism and positional encoding combinations on the  datasets \texttt{GCB}, 
    \texttt{ZINC}, and \texttt{ogbg-molhiv}. Columns indicate combinations of attention mechanisms and positional encodings. Abbreviations for the positional encodings are B for BSPe, H for HodgeLapPE, RB for RWBSPe, T for TopoSlepiansPE, and R for RWPe. Best result, second best result, and third best result for each dataset are highlighted in boldface green, blue, and orange, respectively. 
    For \texttt{GCB},
    the third best result is obtained by the zero positional encoding.}
    \vspace{4pt}
    \label{tab:summary_combinations_attention_pe}
    \begingroup
\begin{tabular}{lccccc}
\toprule
Dataset \hfill $\mathcal{A}_g^s$ 
& B        & H       & RB      & T       & R       \\ 
\midrule\texttt{GCB}  (Accuracy $\uparrow$)         & {\color{ForestGreen}\textbf{0.7516}} & 0.7432  & 0.7442  & 0.7432  & 0.7442  \\
\texttt{ZINC}  (MAE $\downarrow$)        & 0.0840   & {\color{blue}0.0824}& 0.1296  & 0.1303  & 0.1051  \\
\texttt{ogbg-molhiv} (AUC-ROC $\uparrow$)  & {\color{orange}0.7681} & 0.7032  & 0.7082  & 0.7192  & 0.7343  \\
\midrule Dataset \hfill $\mathcal{A}_{k_s\to k_t}^c$                                  
                                  & B        & H       & RB      & T       & R       \\
\midrule\texttt{GCB}  (Accuracy $\uparrow$)                                  & 0.7347   & 0.7337  & 0.7442  & 0.7421  & 0.7379  \\
\texttt{ZINC}  (MAE $\downarrow$)                                  & 0.0833   & {\color{orange}0.0831}& {\color{ForestGreen}\textbf{0.0802}}& 0.1202  & 0.0833  \\
\texttt{ogbg-molhiv} (AUC-ROC $\uparrow$)                                 & {\color{blue}0.7784} & 0.7658  & 0.7321  & 0.7565  & 0.7338  \\
\midrule Dataset \hfill $\mathcal{A}^s_{k_s\to k_t}$
                                  & B        & H       & RB      & T       & R       \\
\midrule\texttt{GCB}  (Accuracy $\uparrow$)                                 & 0.7400   & 0.7421  & 0.7389  & {\color{blue}0.7505}& {\color{blue}0.7505}\\
\texttt{ZINC}  (MAE $\downarrow$)                                  & 0.0934   & 0.0852  & 0.1210  & 0.1452  & 0.0973  \\
\texttt{ogbg-molhiv} (AUC-ROC $\uparrow$)                                   & {\color{ForestGreen}\textbf{0.7946}} & 0.7586  & 0.7058  & 0.7111  & 0.7288  \\
\bottomrule
\end{tabular}
    \endgroup\vspace{-4mm}
\end{table}

Due to the lack of cell complex datasets, as argued in~\autocite{papamarkou2024position}, we test our cellular transformers in three different graph datasets: \texttt{ZINC}~\autocite{zinc_dataset}, \texttt{ogbg-molhiv}~\autocite{ogb_datasets}, and the \texttt{graph classification benchmark} (\texttt{GCB})~\autocite{bianchi2022pyramidal} in its hard version, that we lift to cell complexes. We lift each graph $G$ to a cell complex by filling its cycles with $2$-cells using the \texttt{TopoX} library~\autocite{hajij2024topox}. Cycle filling is not the optimal way of adding cells to our molecules datasets, and may detriment the performance of the transformers. Yet, we will show that it still allows our proposed transformer to achieve results comparable to the state-of-the-art, and thus represents a first step towards encouraging the community to develop cell complex datasets.

For each dataset, we use its official train, validation and test splits, and we try all the possible combinations of the attention mechanisms presented in~\cref{scn:attention_arch} with the positional encodings presented in~\cref{scn:high_order_positional_encodings} plus a positional encoding called zero that assigns a zero vector of fixed length to each cell, simulating absence of positional encodings.
For \texttt{GCB}, we run the experiments with five different random seeds and report average accuracy and standard deviation. Due to computational constraints, we only run the experiments once for the other two datasets \texttt{ZINC} and \texttt{ogbg-molhiv}, reporting only the obtained score. For the state-of-the-art methods, we report the average and standard deviation achieved for these datasets in~\autocite{zhou2024theoretical, bianchi2022pyramidal}. Details on the architectures are in the Appendix.

From state-of-the-art methods in~\autocite{zhou2024theoretical}, we only report the graph and simplicial architectures, and skip the transformers based on clique-lifting, as these cliques do not appear naturally on the graph as high-order cells, and our purpose is developing general cell complex transformers for cell domains. In their experiments, though, the models, while applicable for arbitrary ranks, were tested using only vertex and edge data, excluding higher-order features.
The tensor diagram for the self- and cross-attention and general architectures in each layer can be found in~\cref{fig:td}. Results with our best performing combinations of attention mechanism and positional encodings are 
in~\cref{tab:summary_all_results}. A~summary of the best attention and positional combinations for CT is reported in~\cref{tab:summary_combinations_attention_pe}. Complete experimental results, training details, and hardware resources used for the experiments are in the Appendix.

\subsection{Discussion}
\label{sec:discussion}

Overall, we outperform all previous state-of-the-art methods tested in the \texttt{GCB} dataset and we obtain comparable results to some of the most effective architectures in the other two. 

For \texttt{ZINC}, we surpass most of the message passing architectures with the exception of CIN~\autocite{CWNetworks} and GIN-AK~\autocite{GNNAK}, for which the differences in the performance between them and our models are relatively small compared to the biggest gap in performance for the dataset. For \texttt{ogbg-molhiv}, message passing architectures consistently obtain comparable or better results than graph transformers, except for the GCN~\autocite{GCN} and GIN~\autocite{xu2018powerful} architectures. In the case of graph transformers, the best performing architecture for \texttt{ZINC} is GPS~\autocite{GPS}, which we surpass in \texttt{ogbg-molhiv}. Following GPS, the second best transformer in \texttt{ZINC} is the simplicial transformer of~\autocite{zhou2024theoretical} equipped with virtual simplices, a similar technique to virtual nodes in graph architectures~\autocite{shirzad2023exphormer, hwang2022an}. Without equipping virtual simplices, though, the MAE values for our best model in \texttt{ZINC} and for the simplicial transformers are equal. Both architectures outperform all other non-GPS traditional graph transformers in this dataset, suggesting that high-order interactions are relevant even in the case of graph datasets. For \texttt{ogbg-molhiv}, our best model outperforms graph transformer architectures, and obtains slightly worse performance than the simplicial transformer for vertices and edges of~\autocite{zhou2024theoretical} equipped with virtual simplices. Results without virtual simplices were not reported for \texttt{ogbg-molhiv}.

The results 
corroborate that leveraging high-order information about the dataset in transformer architectures is capable of outperforming or obtaining comparable SOTA results without the need for advanced techniques such as graph rewiring, virtual nodes, or learnable bias matrices in the attention mechanism.

\paragraph{General vs pairwise attention}
We observe that pairwise attention mechanisms obtained the best results for the molecule datasets (\texttt{ZINC} and \texttt{ogbg-molhiv}) and the general attention mechanism obtained the best result for the \texttt{GCB} dataset. 
We observe that in the \texttt{GCB} dataset, vertices, edges, and $2$-cells contain \emph{homogeneous} information about clustering properties of the input graphs, where edges and $2$-cells use concatenations of vertex features associated to the cell. On the other hand, \texttt{ZINC} and \texttt{ogb-molhiv} contain atomic information for the nodes and $2$-cells and bond information for the edges, making the cochains \emph{heterogeneous} at different ranks. The results suggest that
\begin{enumerate}
[topsep=0pt,leftmargin=0.7cm]
\item general attention, using common query and key projections for all cells at the same time, is more suitable for problems where features in all ranks are homogeneous, i.e., 
of the same nature;
\item pairwise attention, that uses specific query and key projections for pairwise ranks, is more suitable for problems where features are heterogeneous, because it allows to each rank to attend to specific properties of each rank in an isolated way.
\end{enumerate}
\paragraph{Global vs local positional encodings}
We classified our positional encodings into two groups depending on whether they were obtained for all ranks simultaneously (global p.e.) or for each rank isolatedly (local p.e.).
We observe that global positional encodings obtained the best results in the three datasets.  However, for \texttt{GCB} and \texttt{ZINC}, the second and third best results used local positional encodings. For \texttt{GCB}, the second place was shared between TopoSlepiansPE and RWPe, while the third place was achieved with the zero positional encoding.
For the \texttt{ZINC} dataset, the second and third places were obtained by HodgeLapPE. Interestingly, neither TopoSlepiansPE nor RWPe obtained good results for the molecule datasets overall. 



\section{Conclusion}
\label{scn:conclusion}
In this work, we introduced the Cellular Transformer (CT), a novel transformer architecture for cell complexes that leverages high-order relationships inherent in topological spaces, together with new positional encodings for it. Our experimental results demonstrated that CT achieves state-of-the-art performance or comparable results without the need for additional enhancements such as virtual nodes or learnable bias matrices. This work motivates further exploration in topological deep learning, particularly in the development of cell complex datasets and the improvement of transformer layers and positional encodings, such as it has happened for graph transformers. Future work will focus on extending the CT framework to more efficient and effective attention mechanisms, exploring more sophisticated positional encodings, and applying the framework to a broader range of real-world datasets and techniques where transformers are being used, such as generative or multi-modal models.

\subsection{Limitations}
\label{scn:limitations}
Cellular transformers suffer from the same computational limitations as traditional and graph transformers~\autocite{vaswani2017attention}, with attention having quadratic complexity on the number of cells.
This means that cellular transformers are currently limited to cell complexes with a low number of cells. However, as in other transformers, computational limitations may be overcome with linearized attention mechanisms~\autocite{choromanski2022rethinking}. We leave the study of efficient cellular transformers as future work. Another drawback that comes with general topological deep learning is that, for harnessing the power of cell complexes in graph datasets, the selection of a good lifting procedure converting graphs into complexes is fundamental. Although there is no general answer as to which lifting algorithm to select in each case, we expect to see proposals in the second edition of the Topological Deep Learning Challenge~\autocite{second_competition_tdl}.

{
\small
 \printbibliography
}


\appendix

\section{Architecture details and experiments} 
\label{scn:architecture_details_and_experiments}

\paragraph{Pairwise attention transformer layer} Following the usual prenorm design~\autocite{on_layer_normalization}, the output of the cellular transformer layer for a specific rank $k_t$ is denoted $\mathbf{X}_{k_t, l+1}$ and computed in six steps, as follows:
\begin{equation*}
    \begin{split}
    &\mathbf{X}^{1}_{k_t, l} = \text{LayerNorm}_{k_t}(\mathbf{X}_{k_t, l}),\\
    &\mathbf{X}^{1}_{k_s, l} = \text{LayerNorm}_{k_s}(\mathbf{X}_{k_s, l})\text{ for each }k_s\text{ in the tensor diagram},\\
    &\mathbf{X}^2_{k_s\to k_t, l} = \mathcal{A}_{k_s\to k_t}^\bullet(\mathbf{X}_{k_t, l}^1, \mathbf{X}_{k_s, l}^1)\text{ for each }k_s\text{ in the tensor diagram},\\
    &\mathbf{X}^3_{k_s\to k_t, l} = \text{Dropout}(\mathbf{X}^2_{k_s\to k_t, l})\text{ for each }k_s\text{ in the tensor diagram},\\
    &\mathbf{X}^4_{k_t, l} = \mathbf{X}_{k_t, l} + \sum_{k_s} \mathbf{X}^3_{k_s\to k_t, l}\\
    &\mathbf{X}^5_{k_t, l} = \text{LayerNorm}(\mathbf{X}^4_{k_t, l}),\\
    &\mathbf{X}^6_{k_t, l} = \text{Dropout}(\text{FFN}_2(\text{Dropout}(\text{ReLU}(\text{FFN}_1(\mathbf{X}^5_{k_t, l}))))),\\
    &\mathbf{X}_{k_t, l+1} = \mathbf{X}^4_{k_t, l} + \mathbf{X}^6_{k_t, l},
    \end{split}
\end{equation*}
The LayerNorm is unique for each dimension $d$ and each layer~$l$. 

\paragraph{General attention transformer layer} Similarly to the pairwise attention transformer layer, the general attention transformer layer introduced in~\cref{scn:general_attention} performs the following steps:
\begin{equation*}
    \begin{split}
    &\mathbf{X}^{1}_{l}= \text{LayerNorm}(\mathbf{X}_{l}),\\
    &\mathbf{X}^2_l= \mathcal{A}_g^{\bullet}(\mathbf{X}_l^1),\\
    &\mathbf{X}^3_l= \text{Dropout}(\mathbf{X}^2_l),\\
    &\mathbf{X}^4_l= \mathbf{X}_{l} + \mathbf{X}^3_l,\\
    &\mathbf{X}^5_l= \text{LayerNorm}(\mathbf{X}^4_l),\\
    &\mathbf{X}^6_l= \text{Dropout}(\text{FFN}_2(\text{Dropout}(\text{ReLU}(\text{FFN}_1(\mathbf{X}^5_l))))),\\
    &\mathbf{X}_{l+1}= \mathbf{X}^4_l + \mathbf{X}^6_l.
    \end{split}
\end{equation*}

\paragraph{Training details}
All experiments use a \texttt{Cosine Annealing} scheduler with linear warmup, an \texttt{AdamW} optimizer with $\epsilon = 1^{-8}$, $(\mu_1, \mu_2) = (0.9, 0.999)$ and variable peak learning rate, and a gradient clipping norm of $5$. All our transformer architectures, after the transformer layers, use a fully-connected readout whose dropout and number of hidden layers is fixed for each set of experiments, followed by a global add pool layer over all the vertex signals to perform prediction or regression. 
The fully-connected block begins with a number of neurons equivalent to the hidden dimension of the transformers and concludes with a number of neurons corresponding to the network's output number. Throughout the block, each hidden layer has half as many neurons as its predecessor.

\paragraph{Architecture details of the cellular transformer}
\Cref{tab:architecture_details_cellular_transformers} presents the hyperparameters of the cellular transformer for the three datasets \texttt{GCB}, \texttt{ogbg-molhiv}, and \texttt{ZINC}.

\begin{table}[htb]
    \centering
    \caption{Cellular transformer parameters for the experiments of the three datasets. The attention type and the positional encodings vary depending on the experiment configuration.}
    \vspace{4pt}
    \label{tab:architecture_details_cellular_transformers}
    \begin{tabular}{lccc}
         \toprule & \texttt{GCB} & \texttt{ogbg-molhiv} & \texttt{ZINC} \\
         \midrule
         \text{\#Layers} & $12$ & $12$ & $12$ \\
         \text{Hidden dimension ($d^h$)} & $80$  & $768$ & $96$ \\
         \text{FFN inner-layer dimension} & $80$ & $768$ & $96$ \\
         \text{\# Attention heads ($m$)} & $8$ & $32$ & $8$ \\
         \text{Hidden dimension of each head} & $10$ & $24$ & $12$ \\
         \text{Attention dropout} & $0.1$ & $0.1$ & $0.1$ \\
         \text{Embedding dropout} & $0.0$ & $0.0$ & $0.0$ \\
         \text{Readout MLP dropout} & $0.1$ & $0.0$ & $0.0$ \\
         \text{Max epochs} & $350$ & $200$ & $10000$\\
         \text{Peak learning rate} & \num{3e-4} & \num{2e-4} & \num{2e-4}\\
         \text{Batch size} & $256$ & $1024$ & $256$\\
         \text{Warmup epochs} & $35$ & $20$ & $1000$\\
         \text{Weight decay} & $0.01$ & \num{1e-5} & $0.01$\\
         \text{\# hidden layers readout MLP} & $1$ & $3$ & $2$\\
         \bottomrule
    \end{tabular}
    
\end{table}

\paragraph{Signals on cells} All graphs in the three datasets contain at least discrete signals for the vertices. For the \texttt{GCB} dataset, we associate to each edge a signal corresponding to concatenating the signals of its endpoints.  For the \texttt{ogbg-molhiv} and \texttt{ZINC} datasets, the edges contain signals, so we do not change them. As a first step in the transformer architecture, we learn an embedding for the discrete features. For the edge features in \texttt{GCB}, each vertex feature is embedded individually. For the three datasets, signals on the $2$-cells are given by sum of the embedded signals of their vertices. 

\paragraph{\texttt{GCB} architectures} The first six models in \cref{tab:summary_all_results} are all graph neural networks with different graph pooling layers and common architecture composition given by MP(32)-Pool-MP(32)-Pool-MP(32)-GlobalPool-Dense(Softmax), where MP(32) is a Chebyshev convolutional layer~\autocite{cheby_conv} with 32 hidden units, Pool is a pooling message passing layer, GlobalPool is a global pool layer used as readout, and Dense(Softmax) is a dense layer with softmax activation. Skip connections were used. The other state-of-the-art models consist of models proposed in~\autocite{graph_embedding_and_classification, hypergraph_kernels_over_sc}.

\section{Mathematical details and examples}
\label{scn:mathematical_details}

\paragraph{Cell complexes}
A definition of cell complexes in the context of algebraic topology can be found in~\autocite{hatcher2005algebraic}. In brief, a cell complex is a topological space $\mathcal{X}$ that can be decomposed as a union of disjoint subspaces called \emph{cells}, where each cell $\sigma$ is homeomorphic to $\mathbb{R}^k$ for some integer $k\ge 0$, called the \emph{rank} of~$\sigma$. Additionally, for every cell~$\sigma$, the difference $\overline{\sigma}\setminus \sigma$ is a union of finitely many cells of lower rank, where $\overline{\sigma}$ denotes the closure of~$\sigma$.
The \emph{dimension} of a finite cell complex is the maximum of the ranks of its cells. The set of cells of rank $k$ in a cell complex $\mathcal{X}$ is denoted by~$\mathcal{X}_k$.
The \emph{$n$-skeleton} of $\mathcal{X}$ is the cell complex spanned by $\mathcal{X}_0,\dots, \mathcal{X}_n$, for $0\le n\le\dim\mathcal{X}$. 

A \emph{characteristic map} for a cell $\sigma$ of rank $k$ is a map from the Euclidean unit closed ball of dimension $k$ into $\overline{\sigma}$ whose restriction to the open ball is a homeomorphism. 
A cell complex is called \emph{regular} if each cell $\sigma$ admits a characteristic map which is itself a homeomorphism from the closed ball to~$\overline{\sigma}$. For example, a decomposition of a circle as the union of a $0$-cell and a $1$-cell is not regular. Geometric realizations of abstract simplicial complexes are regular cell complexes. 
Cell complexes generalize simplicial complexes as their cells are not constrained to be simplices. 

\paragraph{Boundary and coboundary operators}
For each rank~$k$, the incidence matrix $\mathbf{B}_k$ of a cell complex $\mathcal{X}$, as defined in \cref{cc:cell_complexes}, is the matrix of the \emph{boundary operator} $\mathcal{C}_k(\mathcal{X})\to \mathcal{C}_{k-1}(\mathcal{X})$, where $\mathcal{C}_k(\mathcal{X})$ is the $\mathbb{R}$-vector space spanned by the set $\mathcal{X}_k$ of $k$-cells of $\mathcal{X}$. The transpose $\mathbf{B}_k^T$ is the matrix of the \emph{coboundary operator} $\mathcal{C}^{k-1}(\mathcal{X})\to \mathcal{C}^k(\mathcal{X})$ on the dual vector spaces. Thus the matrix $\mathbf{B}_{k}^T$ encodes reverse incidence relations from $(k-1)$-cells to $k$-cells.

\paragraph{Neighborhood matrices}
We jointly call \emph{neighborhood matrices} the (signed or non-signed) incidence matrices, Laplacians, and adjacency matrices.
Laplacians are extensively used in graph learning 
\autocite{benchmarking_gnn, hodge_laplacian_medical} and signal processing \autocite{sardellitti2021topological, Barbarossa_2020}. 
For a cell complex $\mathcal{X}$ of arbitrary dimension, the $k$-th \emph{Hodge Laplacian}
is defined in terms of incidence matrices as
$\mathbf{L}_k=\mathbf{B}_k^T\mathbf{B}_k+\mathbf{B}_{k+1}\mathbf{B}_{k+1}^T$, where $\mathbf{B}_k=0$ if $k=0$ or $k>\dim\mathcal{X}$.
The \emph{upper} and \emph{lower} Laplacians
are, respectively,
$\mathbf{L}_k^{\text{up}}=\mathbf{B}_{k+1}\mathbf{B}_{k+1}^T$ and $\mathbf{L}_k^{\text{down}}=\mathbf{B}_{k}^T\mathbf{B}_{k}$. Therefore, for a 2-dimensional cell complex,
\[
\mathbf{L}_0 = \mathbf{B}_{1}\mathbf{B}_{1}^T,
\qquad
\mathbf{L}_1=
\mathbf{B}_{1}^T\mathbf{B}_{1}+\mathbf{B}_{2}\mathbf{B}_{2}^T,
\qquad
\mathbf{L}_2=
\mathbf{B}_{2}^T\mathbf{B}_{2}.
\]

The lower Laplacian $\mathbf{B}_k^T\mathbf{B}_k$ can be interpreted as the matrix of the composite of the boundary operator $\mathcal{C}_k(\mathcal{X})\to\mathcal{C}_{k-1}(\mathcal{X})$ with the adjoint of the cobundary operator $\mathcal{C}^{k-1}(\mathcal{X})\to \mathcal{C}^k(\mathcal{X})$ through the isomorphism $\mathcal{C}_k(\mathcal{X})\cong\mathbb{R}^{|\mathcal{X}_k|}$ determined by the given order of the set $\mathcal{X}_k$ of $k$-cells of~$\mathcal{X}$.
The upper Laplacian is the composite of the adjoint of the $k$-coboundary with the $(k+1)$-boundary.

Two distinct $k$-cells are upper adjacent if there exists at least one $(k+1)$-cell incident to both. We denote upper adjacency by $\sigma_i\sim_U\sigma_j$. Similarly, two distinct $k$-cells are lower adjacent if they share at least one common incident $(k-1)$-cell. We denote lower adjacency by $\sigma_i\sim_L \sigma_j$ \autocite{Estrada2018Centralities}. The upper and lower adjacency relations are stored using \emph{adjacency matrices} $\textbf{A}^{\text{up}}_k$ and $\textbf{A}^{\text{down}}_k$. These are square matrices of size $|\mathcal{X}_k|=\dim\mathcal{C}^k(\mathcal{X})$, which are related to non-signed incidence matrices as follows: $\textbf{A}^{\text{up}}_k$ is obtained from $\textbf{I}_{k+1}\textbf{I}_{k+1}^T$ by replacing its diagonal entries with zeros, and $\textbf{A}^{\text{down}}_k$ is obtained from $\textbf{I}_{k}^{T}\textbf{I}_{k}$ analogously.


\paragraph{Details on LapPE for graphs and their HodgeLapPE extension}
A popular positional encoding for graph transformers is graph Laplacian eigenvectors (LapPE)~\autocite{benchmarking_gnn}. Let $0\leq \tilde{\lambda}_1 \leq \cdots \leq \tilde{\lambda}_{\tilde p}\leq 2$ be the distinct eigenvalues of the normalized graph Laplacian $\tilde{\mathbf{L}}_0$ for a graph $G=(V,E)$ with $V=\{v_1,\hdots, v_n\}$. As $\tilde{\mathbf{L}}_0$ is a real symmetric matrix,
\begin{equation}
\label{eqn:rayleigh_quotient}
\tilde{\lambda}_i =  \!\!\!\min_{\substack{\lVert x\rVert=1\\ x\in \langle \tilde{V}_{i-1}\rangle^\perp}}
\!\!\! x^T\tilde{\mathbf{L}}_0 x = \!\!\!\min_{\substack{\lVert x\rVert=1\\ x\in \langle \tilde{V}_{i-1}\rangle^\perp}}
\!\!\lVert \mathbf{B}_1^T(D^+)^{1/2}x\rVert^2 = \!\!\!\min_{\substack{\lVert x\rVert=1\\ x\in \langle \tilde{V}_{i-1}\rangle^\perp}}\hspace{0.1cm}\sum_{(v_i, v_j)\in E}\left(\frac{x_i}{\sqrt{d(v_i)}} - \frac{x_j}{\sqrt{d(v_j)}}\right)^2,
\end{equation}
\begin{wrapfigure}[19]{l}{0.3\linewidth}
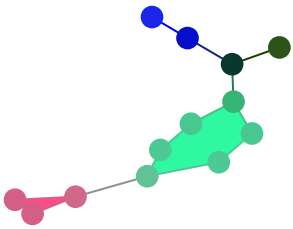

    \centering
    \vspace{-4mm}
    \include{assets/graphBSPeWithoutNumbers.tex}
    \vspace{-2em}
    \caption{BSPe positional encoding of length three for a cell complex with two $2$-cells. To generate a colour from the positional encoding, we normalize each coordinate of the positional encodings to the $[0,1]$ range, generating normalized RGB colours. Note that close cells are assigned similar colours.}
    \label{fig:BSPe_example}
\end{wrapfigure}
where $d(v)$ is the degree of $v$ and $\tilde{V}_{i-1}$ is a set of linearly independent eigenvectors of $\tilde{\lambda}_1, \hdots, \tilde{\lambda}_{i-1}$. The minimum attained for any unit-norm eigenvector of $\tilde{\lambda}_i$ is unique up to a sign for non-degenerate eigenvalues. Eigenvectors corresponding to small eigenvalues give a \textit{gradient} on the graph. The \textit{close} vertices (with respect to the adjacency) have close eigenvector coordinates, thus, represent a \textit{relative} position encoding of the vertices of the graph. Thus, LapPE assigns to each vertex $v_i$ a vector $\text{LapPE}(v_i)=(e^1_{i},\hdots, e^k_{i})$, where $\{e_i^j\mid j=1,\dots,k\}$
are eigenvectors of the 
$k$ smallest eigenvalues counting multiplicities, where $k$ is a hyperparameter. To avoid the sign ambiguity for normalized eigenvectors of non-degenerate eigenvalues, positional encodings are multiplied by random signs at each iteration during training, with the objective of making the neural network invariant to this ambiguity. 

HodgeLapPE is the extension of LapPE to cells of arbitrary rank using the unnormalized Hodge Laplacian. However, without normalizing the Hodge Laplacian, we can obtain eigenvalues that are not small enough to make the previous Rayleigh quotient \eqref{eqn:rayleigh_quotient} meaningful even in the case of simple simplicial complexes. With the unnormalized version we also do not obtain straightforward formulae that require close cells to have close coordinates in the eigenvector. The following examples illustrate the previous drawbacks.

\begin{example}[Large eigenvalues extending LapPE to simplicial complexes using the unnormalized Hodge Laplacian]
Take a triangle graph $G$ 
with vertices $v_1,v_2,v_3$.
The eigenvalues of the unnormalized Hodge Laplacian $\mathbf{L}_1$ given the orientation induced by the ordering of the vertices are $0$ and~$3$.
For $\lambda=3$, an eigenvector of unit length is $(1/\sqrt{2})\,(1, 0 ,1)$,
that assigns the same coordinates to two of the edges and a different coordinate to the other one, although the three edges are clearly neighborhoods and must have similar representations, making eigenvectors useless in this case. 
\end{example}

\begin{example}[Rayleigh quotient of the Hodge Laplacian does not produce a gradient of arbitrary dimensional cells]
If we want to produce positional encodings for $k$-cells using the $k$-th Hodge Laplacian, we obtain
\begin{equation}
\label{eqn:raylegh_hodge_laplacian}
\begin{split}
    \lambda_i &= \min_{\substack{\lVert x\rVert = 1\\ x\in \langle S_{i-1} \rangle^\perp}} x^T \mathbf{L_k} x = \min_{\substack{\lVert x\rVert = 1\\ x\in \langle S_{i-1} \rangle^\perp}}\lVert \mathbf{B}_k x\rVert^2 + \lVert \mathbf{B}_{k+1}^T x\rVert^2\\
    &=\min_{\substack{\lVert x\rVert = 1\\ x\in \langle S_{i-1} \rangle^\perp}}(1-\mathbbm{1}(k=0))\sum_{\gamma\in S_{i-1}}\Bigg(\sum_{\substack{\sigma_j\in S_i\\ \gamma < \sigma_j}}\left(s(\gamma, \sigma_j)x_j\right)\Bigg)^2 \\ & \hspace{2.5cm} + (1-\mathbbm{1}(d=\text{dim}(\mathcal{X})))\sum_{\gamma \in S_{i+1}}\Bigg(\sum_{\substack{\sigma_j\in S_i\\\sigma_j < \gamma}}(s(\sigma_j, \gamma)x_j)\Bigg)^2,
    \end{split}
\end{equation}
where $\gamma < \sigma$ means that $\gamma$ is a proper face of $\sigma$ and $s(\gamma, \sigma)$ is the value of $\gamma$ 
in the boundary of~$\sigma$. In this case, taking the previous triangle graph $G$ with three vertices, the eigenvalue $\lambda=0$ has a unit eigenvector
$(1/\sqrt{3})\,(-1, -1, 1)$, 
that assigns to two of the edges the same coordinates and to the third edge the opposite coordinate, though being the three of them neighbors since they are all adjacent. We refer to~\autocite{Barbarossa_2020} for more information about Equation~\eqref{eqn:raylegh_hodge_laplacian} in the context of Topological Signal Processing.
\end{example}

\section{Positional encoding details}
\label{scn:pe_details_app}

In this section, we extend the details about the positional encodings built in~\cref{scn:high_order_positional_encodings}.

\subsection{Random walks on cell complexes}

Let $\mathcal{X}$ be a regular cell complex. We describe a random walk on the set of $k$-cells of~$\mathcal{X}$. To this end, we first recall that the number of upper and lower adjacent $k$-cells of a given cell $\sigma\in \mathcal{X}_k$ are named respectively the $(0,k+1)$-upper and $(0,k-1)$-lower degree of $\sigma$ \autocite{Hernandez2020SimplicialDegree},
\[
\deg_U^{0,k+1}(\sigma)
=\#\{\sigma'\in \mathcal{X}_{k}\colon \sigma\sim_U\sigma'\}; \qquad 
\deg_L^{0,k-1}(\sigma)=\#\{\sigma'\in \mathcal{X}_{k}\colon \sigma\sim_L\sigma'\}.  
\]
On the one hand, for each $k\geq 0$, we define a random upper $k$-walk based on upper adjacencies of the $k$-cells of $\mathcal{X}$. At each step, we move from a $k$-cell $\sigma_i$ to any upper adjacent $k$-cell $\sigma_j$ with probability proportional to the number of $(k+1)$-cells in common. To describe this process, we consider a weighted undirected graph $G_{k}^{\text{up}}$, whose vertices are the $k$-cells of $\mathcal{X}$ and the weight of each edge $(\sigma_i, \sigma_j)$ is the number of $(k+1)$-cells whose closure contains both cells (if a $k$-cell is not upper adjacent to any $k$-cell, we draw a loop on the corresponding vertex with weight equal to~$1$). Thus, the upper random $k$-walk is described by the left stochastic matrix $\mathbf{RW}_k^{\text{up}}
=\textbf{wA}_k^{\text{up}}(\mathbf{D}^{\text{up}}_k)^{-1}$, where $\textbf{wA}_k^{\text{up}}$ and $\mathbf{D}^{\text{up}}_k$ denote the weighted adjacency and diagonal weighted degree matrices of the graph $G^{\text{up}}_k$. 

On the other hand, for each $k>0$, we define a random lower $k$-walk through lower adjacencies of the $k$-cells of~$\mathcal{X}$. In this case, we move from a $k$-cell $\sigma_i$ to any lower adjacent $k$-cell $\sigma_j$ with probability proportional to the number of $(k-1)$-faces in common. As in the previous case, the random lower walk can be described as a random walk on a weighted graph $G^{\text{down}}_k$, whose vertices are the $k$-cells of $\mathcal{X}$ and the weight of an edge $(\sigma_i,\sigma_j)$ is set as the number of $(k-1)$-cells that both cells have in common (as before, if a $k$-cell is not lower adjacent to any other $k$-cell, then we draw a loop on it with weight equal to $1$). The lower random $k$-walk is described by the left stochastic matrix $\mathbf{RW}_k^{\text{down}}
=\textbf{wA}_k^{\text{down}}(\mathbf{D}^{\text{down}}_k)^{-1}$, where $\textbf{wA}_k^{\text{down}}$ and $\mathbf{D}^{\text{down}}_k$ denote the corresponding weighted adjacency and diagonal weighted degree matrices of the graph $G^{\text{down}}_k$.
The matrices $\textbf{wA}_k^\text{up}$ and $\textbf{wA}_k^\text{down}$ correspond respectively to the upper and lower adjacency matrices $\textbf{A}_k^\text{up}$ and $\textbf{A}_k^\text{down}$ with the diagonal entries in null rows replaced with~$1$.

We can combine both processes to obtain a random walk in which information flows through upper and lower adjacencies, in line with \autocite{high_order_random_walks}. The idea is as follows: if we are in a $k$-cell $\sigma$ with upper and lower adjacent $k$-cells, we take a step with equal probability via either upper or lower connections. If $\sigma$ has upper adjacent $k$-cells but not lower ones, we move following the random upper $k$-walk process, and vice versa. Lastly, if $\sigma$ has neither upper nor lower connections, then we do not move. 

\begin{figure*}[t!]
    \centering
    \begin{subfigure}[t]{0.47\textwidth}
        \centering
        \begin{tikzpicture}

\definecolor{nodecolor}{HTML}{c3dee7}
\definecolor{edgecolor}{HTML}{c3699c}
\definecolor{shapecolor}{HTML}{850a39}

\filldraw[fill=shapecolor, draw=edgecolor, line width=0.8mm] (-1,-0.5) rectangle (1,1); 
\filldraw[fill=shapecolor, draw=edgecolor, line width=0.8mm] (1,1) -- (2, 0) -- (2, 1) -- cycle; 

\draw[->, line width=0.4mm, white] (0, 1) -- (0, 0.2);
\draw[->, line width=0.4mm, black] (0, 1) to[out=90,in=120] (0.9, 1.4);
\draw[->, line width=0.4mm, black] (0, 1) to[out=90,in=60] (-0.9, 1.4);

\foreach \x in {-1, 1} {
    \foreach \y in {-0.5, 1} {
        \fill[nodecolor] (\x,\y) circle (0.25);
        \draw[black] (\x,\y) circle (0.25);
    }
}
\fill[nodecolor] (2, 1) circle (0.25);
\draw[black] (2, 1) circle (0.25);

\fill[nodecolor] (2, 0) circle (0.25);
\draw[black] (2, 0) circle (0.25);

\end{tikzpicture}
        \vspace{-2em}
        \caption{RWBSPe random walk possible transitions from the upper-left edge. }
    \end{subfigure}%
    \hspace{1em}
    \begin{subfigure}[t]{0.47\textwidth}
        \centering
        \begin{tikzpicture}

\definecolor{nodecolor}{HTML}{c3dee7}
\definecolor{edgecolor}{HTML}{c3699c}
\definecolor{shapecolor}{HTML}{850a39}

\filldraw[fill=shapecolor, draw=edgecolor, line width=0.8mm] (-1,-0.5) rectangle (1,1); 
\filldraw[fill=shapecolor, draw=edgecolor, line width=0.8mm] (1,1) -- (2, 0) -- (2, 1) -- cycle; 

\draw[->, line width=0.4mm, black] (0, 1) to[out=280,in=180] (1.3, 0.5);
\draw[->, line width=0.4mm, white] (0, 1) -- (0, -0.35);
\draw[->, line width=0.4mm, white] (0, 1) to[out=-90,in=0] (-0.9, 0.2);
\draw[->, line width=0.4mm, white] (0, 1) to[out=270,in=180] (0.9, 0.2);
\draw[->, line width=0.4mm, black] (0, 1) to[out=90,in=120] (1.5, 1.2);

\foreach \x in {-1, 1} {
    \foreach \y in {-0.5, 1} {
        \fill[nodecolor] (\x,\y) circle (0.25);
        \draw[black] (\x,\y) circle (0.25);
    }
}
\fill[nodecolor] (2, 1) circle (0.25);
\draw[black] (2, 1) circle (0.25);

\fill[nodecolor] (2, 0) circle (0.25);
\draw[black] (2, 0) circle (0.25);

\end{tikzpicture}
        \vspace{-2em}
        \caption{RWPe random walk possible transitions from the upper-left edge.}
    \end{subfigure}
    \caption{Differences between RWBSPe and RWPe random walks. RWBSPe random walks can jump from a cell to all its incident and coincident cells, while 
    RWPe random walks can jump from a cell to all its upper and lower adjacent cells.\vspace{-4mm}}
    \label{fig:rwpe_examples}
\end{figure*}
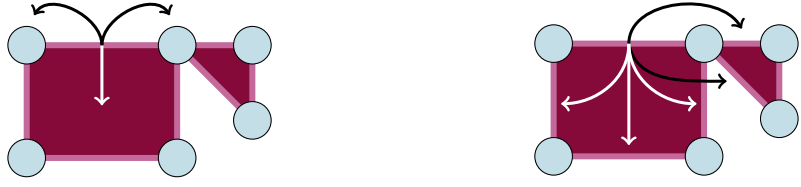

The left stochastic matrix that describes the random $k$-walk is defined for $\sigma_i,\,\sigma_j\in \mathcal{X}_k$ by
\begin{equation*}
    (\mathbf{RW}_k)_{\sigma_i\sigma_j}
    =\begin{cases}
    \frac{1}{2}(\mathbf{RW}^{\text{up}}_k)_{\sigma_i\sigma_j}+\frac{1}{2}(\mathbf{RW}^{\text{down}}_k)_{\sigma_i\sigma_j} & \text{ if } \deg_U^{0,k+1}(\sigma_j)\neq 0\text{ and } \deg_L^{0,k-1}(\sigma_j)\neq 0 \\
    (\mathbf{RW}^{\text{up}}_k)_{\sigma_i\sigma_j} & \text{ if }\deg_U^{0,k+1}(\sigma_j)\neq 0\text{ and } \deg_L^{0,k-1}(\sigma_j)= 0 \\
    (\mathbf{RW}^{\text{down}}_k)_{\sigma_i\sigma_j} & \text{ if }\deg_U^{0,k+1}(\sigma_j)=0\text{ and } \deg_L^{0,k-1}(\sigma_j)\neq 0\\
    \mathbbm{1}(i=j) & \text{ if }\deg_U^{0,k+1}(\sigma_j)=\deg_L^{(0,k-1)}(\sigma_j)=0.
\end{cases}
\end{equation*}
An example of the differences between transitions from an edge in the random walks described in this section and the barycentric subdivision random walks of RWBSPe are described in~\cref{fig:rwpe_examples}.

\subsection{Topological Slepians}
Let $\mathcal{X}$ be an oriented regular two-dimensional cell complex, with Hodge Laplacians $\mathbf{L}_1$ and $\mathbf{L}_2$.  Hodge Laplacians admit a Hodge decomposition \cite{lim2020hodge}, such that the $k$-cochain space can be decomposed as
\begin{equation} \label{hodge_spaces}
\mathcal{C}^{k}(\mathcal{X},\mathbb{R}) = \text{im}\big(\mathbf{B}_{k}^T\big) \oplus \text{im}\big(\mathbf{B}_{k+1}\big) \oplus \text{ker}\big(\mathbf{L}_{k}\big),
\end{equation}
where $\bigoplus$ denotes direct sum of vector spaces, and $\text{ker}(-)$ and $\text{im}(-)$ are the kernel and image spaces of a matrix, respectively. The $k$-cochains can be represented by means of eigenvector bases of the corresponding Hodge Laplacian. Using the 
decomposition
$\mathbf{L}_k=\mathbf{U}_k \mathbf{\Lambda}_k\mathbf{U}_k^T$, the $k$-th Cellular Fourier Transform ($k$-CWFT) is the projection of a $k$-cochain onto the eigenvectors of $\mathbf{L}_k$ \cite{sardellitti2022topological}:
\begin{equation}
\label{k-GFT}
\widehat{\mathbf{X}}_k = \mathbf{U}_k^T\, \mathbf{X}_k.
\end{equation}
We refer to the eigenvalue set $\mathcal{B}_k$ of the $k$-CWFT as the frequency domain. An immediate consequence of the Hodge decomposition in (\ref{hodge_spaces}) is that the eigenvectors belonging to $\text{im}(\mathbf{L}^{d}_k)$ are orthogonal to those belonging to  $\text{im}(\mathbf{L}^{u}_k)$, for all $k=1,\ldots,K-1$. Therefore, the eigenvectors of $\mathbf{L}_k$ are given by the union of the eigenvectors of $\mathbf{L}_k^u$, the eigenvectors of $\mathbf{L}_k^d$, and the kernel of $\mathbf{L}_k$. We now introduce two localization operators acting onto a $k$-cell concentration set (thus, onto the topological domain), say $\mathcal{S}_k\subset \mathcal{X}^k$, and onto a spectral concentration set (thus, onto the frequency domain), say $\mathcal{F}_k\subset \mathcal{B}_k$, respectively. In particular, we  define a cell-limiting operator onto the $k$-cell set $\mathcal{S}_k$ as 
\begin{align}\label{edge_limiting_operator}
\mathbf{C}_{\mathcal{S}_k}={\rm diag}(\mathbf{1}_{\mathcal{S}}) \in\mathbb{R}^{|\mathcal{X}_k|\times |\mathcal{X}_k|} ,
\end{align}
where $\mathbf{1}_{\mathcal{S}_k}\in\mathbb{R}^{|\mathcal{X}_k|}$ is a vector having ones in the index positions specified in $\mathcal{S}_k$, and zero otherwise; and ${\rm diag}(\mathbf{z})$ denotes a diagonal matrix having $\mathbf{z}$ on the diagonal. A~$k$-cochain $\mathbf{X}_k$ is perfectly localized onto the set $\mathcal{S}_k$ if $\mathbf{C}_{\mathcal{S}_k} \mathbf{X}_k=\mathbf{X}_k$. Similarly, the frequency limiting operator is defined as
\begin{align}\label{frequency_limiting_operator}
\mathbf{B}_{\mathcal{F}_k}=\mathbf{U}\,{\rm diag}(\mathbf{1}_{\mathcal{F}_k})\,\mathbf{U}^T \in\mathbb{R}^{|\mathcal{X}_k|\times |\mathcal{X}_k|},
\end{align}
that can be interpreted as a band-pass filter over the frequency set $\mathcal{F}_k$. A $k$-cochain is perfectly localized over the bandwidth $\mathcal{F}_k$ if $\mathbf{B}_{\mathcal{F}_k} \mathbf{X}_k=\mathbf{X}_k$. 
The matrices in (\ref{edge_limiting_operator}) and (\ref{frequency_limiting_operator}) are proper projection operators.

At this point, \emph{$k$-topological Slepians} are defined as orthonormal vectors that are maximally concentrated over the $k$-cell set $\mathcal{S}_k$, and perfectly localized onto the bandwidth $\mathcal{F}_k$:
\begin{align}
    s_k^i = &\argmax_{s_k^i}\; \lVert\mathbf{C}_{\mathcal{S}_k} s_k^i\rVert_2^2 \nonumber \\
    &\textrm{ subject to} \;\; \label{slep_prob}   \lVert s_k^i\rVert = 1, \quad \mathbf{B}_{\mathcal{F}_k} s_k^i = s_k^i, \\
    & \; \qquad \qquad \langle s_k^i,s_k^j\rangle = 0, \quad \hbox{$j = 1,\ldots,i-1$, if  $i > 1$,} \nonumber
\end{align}
for $i=1,\ldots,|\mathcal{X}_k|$. As shown in \cite{topo_slepians}, the solution of problem \eqref{slep_prob} is given by the eigenvectors of the matrix operator $\mathbf{B}_\mathcal{F}\mathbf{C}_\mathcal{F}\mathbf{B}_{\mathcal{F}_k}$, i.e.,
\begin{equation} \label{slep_sol}
    \mathbf{B}_{\mathcal{F}_k}\mathbf{C}_{\mathcal{S}_k}\mathbf{B}_{\mathcal{F}_k}s_k^i = \lambda_k^i s_k^i.
\end{equation}
It is then clear that the maximum number of $k$-topological Slepians per each pair of concentration sets $\{\mathcal{S}_k,\mathcal{F}_k\}$ is given by $\textrm{rank}\{\mathbf{B}_{\mathcal{F}_k}\mathbf{C}_{\mathcal{S}_k}\mathbf{B}_{\mathcal{F}_k}\}$. For this reason and to have a more exhaustive representation of structural and topological properties of the complex \cite{topo_slepians}, we choose a sequence of $M$ concentration sets $\{\mathcal{S}_{k,i},\mathcal{F}_{k,i}\}_{i=1}^M$ and concatenate the corresponding Slepians. From (\ref{hodge_spaces}), the frequency domain can be partitioned into two separate sets: (i) the set of eigenvalues of $\mathbf{L}_k^{\text{down}}$, say $\mathcal{F}_k^d$, and (ii) the set of eigenvalues of $\mathbf{L}_k^{\text{up}}$, say $\mathcal{F}_k^u$. For the same reason,  we can define two distinct sequences of \textcolor{black}{(not necessarily disjoint)} $k$-cell concentration sets: (i) $K_k^d$ sets based on lower adjacency (encoded by~$\mathbf{L}_k^{\text{down}}$), that we refer to as \emph{lower sets}; (ii)  $K_k^u$ sets based on upper adjacency (encoded by~$\mathbf{L}_k^{\text{up}}$), that we refer to as \emph{upper sets}. It is then natural to associate the frequency concentration set $\mathcal{F}_k^u$ to each upper $k$-cell concentration set, and the frequency concentration set $\mathcal{F}_k^d$ to each lower $k$-cell concentration set. Finally, suppose that the kernel of the Laplacian $\mathbf{L}_k$ is not empty. In that case, topological Slepians can be combined with the harmonic eigenvectors of $\mathbf{L}_k$, i.e. the eigenvectors associated with the zero eigenvalues, resulting in a mixed positional encoding strategy. The number of topological Slepians can then be controlled either by tuning $K_k^d$ and $K_k^u$, or by taking just the top Slepians per each pair of concentration sets.  In this work, we choose the upper and lower $k$-cell concentration sets as the adjacency and coadjacency of each $k$-cell including the $k$-cell itself, respectively, obtaining $|\mathcal{X}_k|$ pairs of concentration sets for each rank~$k$. In this way, we are also sure that the obtained set of topological Slepians comes with theoretical guarantees, i.e., it is an $(A,B)$-frame \cite{vetterlispbook,topo_slepians}  In general, the choice of the localization sets is an interesting directions, that could be easily combined with prior knowledge about the complex and the data at hand.

\begin{table}[h!]
    \centering
    \caption{Results of experiments with cellular transformers in the three datasets \texttt{GCB}, \texttt{ogbg-molhiv}, and
        \texttt{ZINC}. Reported scores are test accuracy ($\uparrow$), test AUC-ROC ($\uparrow$), and test MAE ($\downarrow$), respectively.
        Best result, second best result, and third best result are highlighted in boldface green, blue, and orange, respectively.}
    \vspace{4pt}
    \label{tab:full_experiment_results}
    \begin{tabular}{ccccc}
        \toprule
        \multicolumn{2}{c}{Datasets $\to$} & \texttt{GCB}    & \texttt{ogbg-molhiv}                & \texttt{ZINC}                                    \\
        \midrule
        Attention type                     & Positional encodings & Accuracy ($\uparrow$)               & AUC-ROC ($\uparrow$)   & MAE ($\downarrow$)      \\
        \midrule
        $\mathcal{A}_{g}^s$                & BSPe                 & {\color{ForestGreen}\textbf{0.7516 $\pm$ 0.0102}}    & {\color{orange}0.7681} & 0.0840                  \\
        $\mathcal{A}_{k_s\to k_t}^s$       & BSPe                 & 0.7400 $\pm$ 0.0123                 & {\color{ForestGreen}\textbf{0.7946}}    &   0.0934                      \\
        $\mathcal{A}_{k_s \to k_t}^d$      & RWBSPe               & 0.6295 $\pm$ 0.0263                 & 0.7136                 & 0.3451                  \\
        $\mathcal{A}_{g}^s$                & zeros                & 0.7453 $\pm$ 0.0086                 & 0.7362                 & 0.1239                  \\
        $\mathcal{A}_{k_s \to k_t}^c$      & HodgeLapPE          & 0.7337 $\pm$ 0.0103                 & 0.7658                 & {\color{orange}0.0831} \\
        $\mathcal{A}_{k_s \to k_t}^d$      & HodgeLapPE          & 0.6179 $\pm$ 0.0286                 & 0.7670                 & 0.4009                  \\
        $\mathcal{A}_{k_s\to k_t}^s$       & zeros                & 0.7453 $\pm$ 0.0123                 & 0.7008                 & 0.1416                  \\
        $\mathcal{A}_{k_s\to k_t}^s$       & HodgeLapPE          & 0.7421 $\pm$ 0.0191                 & 0.7586                 & 0.0852                  \\
        $\mathcal{A}_{g}^d$                & BSPe                 & 0.5989 $\pm$ 0.0219                 & 0.7177                 & 0.3983                  \\
        $\mathcal{A}_{g}^s$                & RWBSPe               & 0.7442 $\pm$ 0.0140                 & 0.7082                 & 0.1296                  \\
        $\mathcal{A}_{k_s \to k_t}^c$      & zeros                & {\color{orange}0.7484 $\pm$ 0.0061} & 0.7435                 & 0.1090                  \\
        $\mathcal{A}_{k_s \to k_t}^d$      & BSPe                 & 0.6095 $\pm$ 0.0234                 & 0.7328                 & 0.4096                  \\
        $\mathcal{A}_{k_s \to k_t}^d$      & zeros                & 0.6400 $\pm$ 0.0196                 & 0.5952                 & 0.4348                  \\
        $\mathcal{A}_{k_s \to k_t}^c$      & TopoSlepiansPE       & 0.7421 $\pm$ 0.0094                 & 0.7565                 & 0.1202                  \\
        $\mathcal{A}_{k_s \to k_t}^c$      & RWPe                 & 0.7379 $\pm$ 0.0102                 & 0.7338                 & 0.0833                  \\
        $\mathcal{A}_{g}^d$                & RWBSPe               & 0.6200 $\pm$ 0.0423                 & 0.6949                 & 0.3586                  \\
        $\mathcal{A}_{g}^d$                & TopoSlepiansPE       & 0.6179 $\pm$ 0.0276                 & 0.6954                 & 0.5175                  \\
        $\mathcal{A}_{k_s \to k_t}^c$      & BSPe                 & 0.7347 $\pm$ 0.0136                 & {\color{blue}0.7784}   & 0.0833                  \\
        $\mathcal{A}_{k_s \to k_t}^c$      & RWBSPe               & 0.7442 $\pm$ 0.0108                 & 0.7321                 & {\color{ForestGreen}\textbf{0.0802}}    \\
        $\mathcal{A}_{g}^s$                & HodgeLapPE          & 0.7432 $\pm$ 0.0112                 & 0.7032                 & {\color{blue}0.0824}   \\
        $\mathcal{A}_{g}^s$                & RWPe                 & 0.7442 $\pm$ 0.0136                 & 0.7343                 & 0.1051                  \\
        $\mathcal{A}_{g}^d$                & zeros                & 0.6105 $\pm$ 0.0221                 & 0.6679                 & 0.4526                  \\
        $\mathcal{A}_{k_s \to k_t}^d$      & RWPe                 & 0.6463 $\pm$ 0.0406                 & 0.6944                 & 0.3843                  \\
        $\mathcal{A}_{k_s\to k_t}^s$       & RWBSPe               & 0.7389 $\pm$ 0.0144                 & 0.7058                 & 0.1210                  \\
        $\mathcal{A}_{k_s \to k_t}^d$      & TopoSlepiansPE       & 0.6463 $\pm$ 0.0214                 & 0.6690                 & 0.4527                  \\
        $\mathcal{A}_{g}^d$                & HodgeLapPE          & 0.5811 $\pm$ 0.0250                 & 0.7243                 & 0.3981                  \\
        $\mathcal{A}_{k_s\to k_t}^s$       & TopoSlepiansPE       & {\color{blue}0.7505 $\pm$ 0.0054}   & 0.7111                 & 0.1452                  \\
        $\mathcal{A}_{g}^s$                & TopoSlepiansPE       & 0.7432 $\pm$ 0.0107                 & 0.7192                 & 0.1303                  \\
        $\mathcal{A}_{k_s\to k_t}^s$       & RWPe                 & {\color{blue}0.7505 $\pm$ 0.0054}   & 0.7288                 & 0.0973                  \\
        $\mathcal{A}_{g}^d$                & RWPe                 & 0.6368 $\pm$ 0.0282                 & 0.6836                 & 0.3770                  \\
        \bottomrule
    \end{tabular}
\end{table}

\section{Numerical results}
The set of results of our experiments for all the combinations of attention mechanisms and positional encodings is reported in~\cref{tab:full_experiment_results}.

\section{Implementation and hardware resources}
\label{scn:implementation_and_hardware_resources}

Implementation was performed mainly using the \texttt{TopoNetX}~\autocite{toponetx} library for cell complex representation and manipulation, \texttt{PyTorch}~\autocite{pytorch} for the deep learning pipelines, \texttt{PyTorch Geometric}~\autocite{pytorch_geometric} for feature pooling and dataset loading, \texttt{Deep Graph Library}~\autocite{wang2019dgl} and \texttt{Scipy}~\autocite{scipy} for sparse tensor algebraic operations and sparse tensor representation and manipulation, \texttt{NetworkX}~\autocite{networkx} for graph manipulation, and \texttt{PyTorch Lightning}~\autocite{Falcon_PyTorch_Lightning_2019} as a top layer for experimentation in \texttt{PyTorch}. The most critical pieces of software implemented in this project have been the \texttt{DataLoader} and the \texttt{collate} function to batch cell complexes. The \texttt{DataLoader} is implemented in the class \texttt{TopologicalTransformerDataLoader}. The collate function is implemented in the function \texttt{collate}, both inside the file \texttt{src/datasets/cell\_dataloader.py}. The \texttt{collate} function creates a cell complex batch by performing the disjoint union of cell complexes. As an input, the \texttt{collate} function receives an object with the signals for each cell, the neighborhood matrices used in the transformer architecture as bias $\mathbf{N}$ in a sparse format, and other data needed by the experiments such as the label of the dataset and the positional encodings. The neighborhood matrices are batched into a new sparse block matrix, taking into account that different cell complexes may have different dimensions and thus not all the cell complexes have the same neighborhood matrices. Currently, the \texttt{collate} function supports adjacency and boundary matrices, although the function can be extended easily. Signals, positional encodings, and labels are simply concatenated. To keep track of which signals and positional encodings correspond to each of the individual cell complex, we also return, for each dimension, a vector of size equal to total number of cells of that dimension in the disjoint union which indicates to which cell complex belong each signal or positional encoding.

The experiments were executed on a server with an AMD EPYC 7452 (128) @ 2.350GHz CPU, 503GiB of RAM memory, x4 PNY Nvidia RTX 6000 Ada Generation 48GB GPUs, and Ubuntu 22.04.4 LTS with the 6.5.0-28-generic Linux kernel. Each experiment was executed on a separated GPU device, using 12 workers per experiment.

\section{Licenses}
\label{scn:licenses}

The \texttt{GCB} dataset is distributed under a MIT license. \texttt{ogbg-molhiv} is distributed under a MIT license. The \texttt{ZINC} dataset is free to use and download and its license can be found at~\url{https://wiki.docking.org/index.php?title=UCSF_ZINC_License}. 

\end{document}